\documentclass[final]{cvpr}

\usepackage{algorithm}
\usepackage{amsmath}
\usepackage{amssymb}
\usepackage[toc,page]{appendix}
\usepackage{booktabs}
\usepackage{color}
\usepackage{colortbl}
\usepackage{comment}
\usepackage{enumitem}
\usepackage{epsfig}
\usepackage{etoolbox}
\usepackage{graphicx}
\usepackage[utf8]{inputenc}
\usepackage{listings}
\usepackage{multirow}
\usepackage[numbers,sort&compress]{natbib}
\usepackage[olditem,oldenum]{paralist}
\usepackage{pifont}
\usepackage{tabularx}
\usepackage{times}

\newcommand{\myapprox}{{\raise.17ex\hbox{$\scriptstyle\sim$}}}

\newcommand{\reffig}[1]{Fig.~\ref{#1}}
\newcommand{\refsec}[1]{Sec.~\ref{#1}}

\usepackage{transparent}
\usepackage[dvipsnames]{xcolor}
\usepackage{makecell}
\usepackage{subcaption}

\usepackage[pagebackref,breaklinks,colorlinks,bookmarks=False]{hyperref}
\usepackage{cleveref}

\usepackage{inconsolata}

\hypersetup{
   urlcolor=blue,
   citecolor=ForestGreen,
}

\def\cvprPaperID{399}
\def\confYear{CVPR 2021}
\usepackage{array}
\newcolumntype{P}[1]{>{\centering\arraybackslash}p{#1}}

\newcommand{\app}{CAST}

\newcommand{\fullapp}{Contrastive Attention-Supervised Tuning}

\definecolor{orange}{rgb}{1,0.5,0}
\definecolor{lightsalmonpink}{rgb}{1.0, 0.6, 0.6}
\definecolor{verylightsalmonpink}{rgb}{0.966, 0.805, 0.797}
\definecolor{lightblue}{rgb}{0.862, 0.906, 0.984}
\definecolor{lightyellow}{rgb}{1.0, 0.945, 0.797}
\definecolor{lightgreen}{rgb}{0.835, 0.91, 0.828}
\definecolor{lightpurple}{rgb}{0.879, 0.832, 0.902}

\newcommand{\imagenet}[0]{ImageNet-1k}
\newcommand{\voc}[0]{PASCAL VOC}
\newcommand{\inclf}[0]{IN-1k}
\newcommand{\vocclf}[0]{VOC07}

\newcommand{\random}[0]{Random Init}
\newcommand{\virtex}[0]{VirTex}
\newcommand{\moco}[0]{MoCo}

\newcommand{\mocococo}[0]{MoCo-COCO}

\newcommand{\ttbf}[1]{\textbf{\texttt{#1}}}
\newcommand{\graycell}{\cellcolor{gray!20}}
\newcommand{\band}{\rowcolor{gray!20}}
\newcommand{\drop}[1]{\textcolor{gray}{\textsubscript{$-$#1}}}
\newcommand{\rise}[1]{\textcolor{gray}{\textsubscript{$+$#1}}}
\newcommand{\riseneg}[1]{\textcolor{gray}{\textsubscript{$-$#1}}}

\newcommand{\Drop}[1]{\textcolor{Red}{\textsubscript{\bf $-$#1}}}
\newcommand{\Rise}[1]{\textcolor{Green}{\textsubscript{\bf $+$#1}}}
\newcommand{\Riseneg}[1]{\textcolor{Green}{\textsubscript{\bf $-$#1}}}

\belowdisplayskip \abovedisplayskip

\newlength{\abstractReduceTop}
\newlength{\abstractReduceBot}

\newcommand{\reducedSection}[1]{\vspace{\sectionReduceTop}\section{#1}\vspace{\sectionReduceBot}}
\newlength{\sectionReduceTop}
\newlength{\sectionReduceBot}

\newlength{\subsectionReduceTop}
\newlength{\subsectionReduceBot}

\newlength{\subsubsectionReduceTop}
\newlength{\subsubsectionReduceBot}

\newlength{\captionReduceTop}
\newlength{\captionReduceBot}

\newlength{\eqnReduceTop}
\newlength{\eqnReduceBot}

\newlength{\horSkip}
\newlength{\verSkip}

\newlength{\figureHeight}
\begin{document}

\title{CASTing Your Model: \\Learning to Localize Improves Self-Supervised Representations\vspace{-10pt}}


\author{Ramprasaath R. Selvaraju$^1$\thanks{Equal Contribution} \and
Karan Desai$^{2*}$ \and 
Justin Johnson$^2$ \and 
Nikhil Naik$^1$ \and \vspace{-10pt}\\
$^1$Salesforce Research, $^2$University of Michigan \\
{\tt\small \{rselvaraju,nnaik\}@salesforce.com}
{\tt\small \{kdexd,justincj\}@umich.edu}
}
\maketitle

\thispagestyle{plain}
\pagestyle{plain}

\begin{abstract}
Recent advances in self-supervised learning (SSL) have largely closed the gap with supervised ImageNet pretraining.
Despite their success these methods have been primarily applied to unlabeled ImageNet images, and show marginal gains when trained on larger sets of uncurated images.
We hypothesize that current SSL methods perform best on iconic images, and struggle on complex scene images with many objects.
Analyzing contrastive SSL methods shows that they have poor visual grounding and receive poor supervisory signal when trained on scene images.
We propose Contrastive Attention-Supervised Tuning (CAST) to overcome these limitations.
CAST uses unsupervised saliency maps to intelligently sample crops, and to provide grounding supervision via a Grad-CAM attention loss.
Experiments on COCO show that CAST significantly improves the features learned by SSL methods on scene images, and further experiments show that CAST-trained models are more robust to changes in backgrounds.

\end{abstract}

\vspace{-5pt}
\reducedSection{Introduction}


{
 {
   Self-supervised learning} (SSL) of visual feature representations has seen great interest in recent years.
 SSL in computer vision aims to learn feature representations without using any human annotations, which can be utilized by downstream tasks such as supervised image classification~\cite{krizhevsky2012imagenet,russakovsky2015imagenet}, object detection~\cite{girshick2014rich,ren2015faster}, and semantic segmentation~\cite{long2015fully,he2017mask}.
 Recent SSL methods based on contrastive learning~\cite{gutmann2010nce,hadsell2006dimensionality} have begun to match or even outperform supervised pretraining on several downstream tasks~\cite{wu2018npid,chen2020simclr,misra2019pirl,he2019moco,junnan2020pcl,caron2020swav}.
}



The promise of self-supervised methods is that they ought to allow us to learn better features by scaling to ever-larger training sets, without the need for expensive human-provided labels.
Unfortunately, the success of recent SSL methods has been largely confined to unlabeled images from the ImageNet~\cite{russakovsky2015imagenet} training set.
Na\"{i}vely applying them to larger uncurated sets of internet images has shown marginal gains~\cite{he2019moco,misra2019pirl,caron2020swav} despite using image sets that are orders of magnitude larger than ImageNet (eg. Instagram-1B~\cite{mahajan2018exploring}, YFCC100M~\cite{yfcc100m}, JFT-300M~\cite{sun2017revisiting}).

We hypothesize that current SSL methods perform best when trained on \emph{iconic images} of single objects (like those in ImageNet) but struggle when trained on more complex \emph{scene images} with many objects.
Indeed, current SSL methods struggle even when trained on curated datasets of scene images~\cite{desai2020virtex,grill2020bootstrap} such as COCO~\cite{lin2014microsoft} or Places205~\cite{zhou2014places}.

\begin{figure}[t]
    \begin{subfigure}[t]{\columnwidth}
    \includegraphics[scale=0.124]{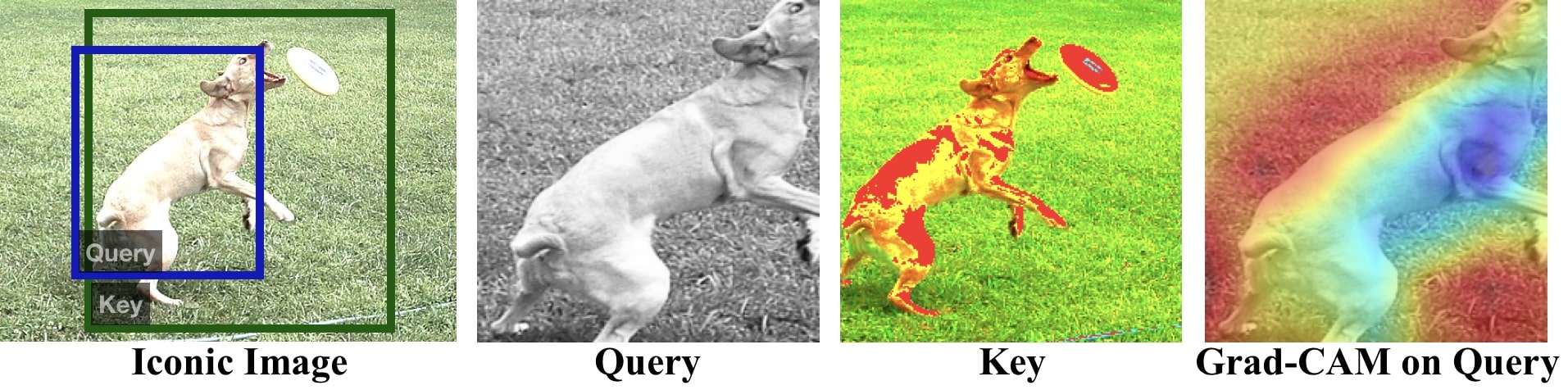}
    \vspace{-12pt}
    \caption{Poor visual grounding ability}
    \label{fig:poor_vis_grounding}
    \end{subfigure}
    \begin{subfigure}[t]{\columnwidth}
    \includegraphics[scale=0.124]{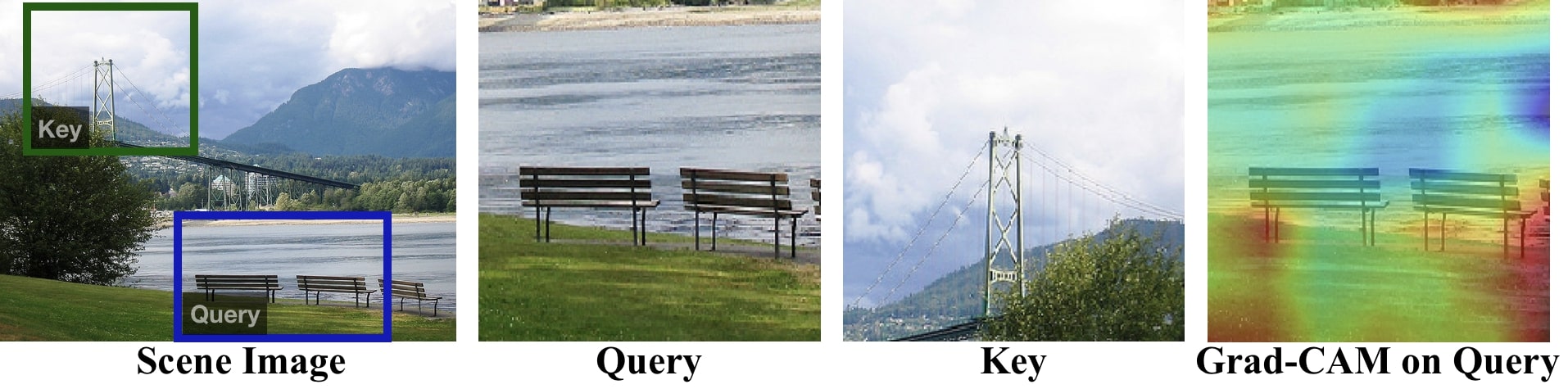}
    \caption{Sampling issues with complex images}
    \vspace{-5pt}
    \label{fig:scene_level_images}
    \end{subfigure}
    \caption{
      We identify two issues with recent contrastive approaches to self-supervised learning:
      (a) \textbf{Poor grounding:} On iconic images, contrastive methods can match key and query but use the wrong image regions to do so. Grad-CAM~\cite{selvaraju2017gradcam} reveals that the model puts high weight (red) on background regions, and low weight (blue) on the object of interest.
      (b) \textbf{Inconsistent Samples:} On complex images, randomly sampled crops may portray different objects, giving an inconsistent learning signal.
      We show that correcting these issues improves self-supervised learning.
    }
	\label{fig:problems_with_ssl}
	\vspace{-15pt}
\end{figure}


In this paper, we analyze contrastive self-supervised models to understand the cause of these limitations and propose a solution to overcome them. Specifically, we find that existing contrastive self-supervised models have poor visual grounding ability and they  receive imperfect supervisory signal when augmented views contain different visual concepts, which is common in images of complex scenes. 

{These issues may arise from  the practice of training the instance discrimination task with random views from images. This practice does not encourage semantic understanding, and models often {cheat} by exploiting low-level visual cues or spurious background correlations.  For example, in \Cref{fig:poor_vis_grounding}, the model relies on the grass to match the two augmented views of the dog. 
Augmented views for training these models commonly start with taking random crops from an image. This strategy may be acceptable for iconic images. However, for scene images, like those in COCO, two views may contain semantically distinct objects (such as the crops in \Cref{fig:scene_level_images}).
This fact may explain diminishing improvements of contrastive SSL models trained on varied web images, and the reduction in their performance when trained with scene images alone. 
}

To mitigate these limitations, we propose Contrastive Attention-Supervised Tuning (\app{}), a training method to improve the visual grounding ability of contrastive SSL methods. 
CAST consists of two algorithmic components: (a) an intelligent geometric transform for cropping different views from an input image, based on constraints derived from an unsupervised saliency map, and (b) a Grad-CAM~\cite{selvaraju2017gradcam}-based attention loss  that provides explicit grounding supervision by forcing the model to attend to objects that are common across the  crops. 

We train the Momentum Contrastive Encoder (MoCo)~\cite{he2019moco}, a leading contrastive learning method, using CAST on the COCO dataset.  We evaluate its performance using image classification, object detection, and instance segmentation tasks, obtaining robust gains in all cases. Additional experiments on the Backgrounds Challenge~\cite{xiao2020noise} show that CAST-trained models are substantially more resilient to changes in object backgrounds when performing image classification. Finally, qualitative and quantitative experiments show that CAST improves object localization ability of contrastive SSL feature representations on COCO scene images and on downstream image classification tasks. We hope that CAST can enable self-supervised learning from unconstrained web-scale  datasets containing images with complex interactions of multiple objects and lead to better out-of-distribution performance and greater robustness to contextual bias.

\reducedSection{Related Work}
\label{sec:rel}
\paragraph{Self-supervised learning:} SSL methods learn features from unlabeled data  using ``pretext'' tasks that provide free supervision, with the aim of performing well on related supervised learning tasks. A strand of research includes low- to high-level computer vision-based pretext tasks, including image inpainting~\cite{Pathak_2016_inpaiting}, colorization~\cite{Zhang_2016_color,Zhang_2017_splitbrain}, predicting patch orderings~\cite{Doersch_2015_context,Noroozi_2016_jigsaw} or degree of rotation~\cite{gidaris2018rotnet}. Pretext tasks that perform pseudo-labeling and clustering~\cite{caron2018deepcluster,caron2019unsupervised,asano2019self,caron2020swav,junnan2020pcl,grill2020bootstrap} have also been shown to be effective. Recently, contrastive learning methods~\cite{hadsell2006dimensionality} that learn to perform instance discrimination~\cite{Wu_2018_Instance,Ye_2019_end2end,oord2018cpcv1,tian2019cmc,he2019moco,chen2020simclr,chen2020big,chen2020improved} have been shown to be the most competitive with fully supervised learning. 

As a result, recent work has focused on developing theoretical and empirical understanding of contrastive representations~\cite{tian2020makes,wang2020understanding,tian2020understanding} and improving the learning framework. For instance, Purushwalkam and Gupta~\cite{purushwalkam2020demystifying} propose a method to improve viewpoint invariance of contrastively-learnt representations . Zhang and Maire~\cite{zhang2020self} utilize a hierarchical region structure of images to guide contrastive learning methods for improved segmentation performance.  In this paper, we show the utility of visual grounding for improving the contrastive representation learning. \vspace{-10pt}

\paragraph{Visual Grounding and Attention:} Improving visual grounding of CNNs is an increasingly important computer vision problem , which can benefit applications such as image captioning~\cite{ma2020learning}, visual question answering~\cite{antol2015vqa}, and debiased computer vision~\cite{jia2018right}. Grounding methods in these problems typically use human attention supervision~\cite{liu2016attention,Selvaraju_2019_ICCV,qiao2018exploring}. In our work, we improve visual grounding of self-supervised models using object saliency maps.\vspace{-10pt}

\paragraph{Object Saliency Prediction:} The goal of object saliency prediction is to identify and segment important objects of interest in an image. 
Saliency prediction methods can be classified into supervised and unsupervised methods. 
Supervised methods for saliency prediction~\cite{jiang2013salient,zhu2014saliency} typically rely on expensive human annotated training data.   
Classically, unsupervised saliency prediction methods utilized handcrafted priors based on human perception~\cite{bruce2005saliency,harel2007graph} or image statistics~\cite{goferman2011context,cheng2014global,zhu2014saliency}. More recently proposed neural network-based saliency prediction methods~\cite{zhang2017supervision,zhang2018deep} utilize saliency maps from the handcrafted unsupervised methods as noisy psuedo-labels for training, thus removing the need for  human-labeled data. In this work we make use of Deep-USPS~\cite{nguyen2019deepusps}, an unsupervised saliency prediction algorithm which uses a two stage mechanism that combines hand-crafted supervision and iterative self-training. 






\section{\fullapp{}}
\label{sec:preliminaries}






Our method, which we call \fullapp{} (\app{}), aims to tune self-supervised models to encourage them to rely on the appropriate regions during contrastive learning. At a high level, \app{} consists of two steps: 1. constrained sampling of the query and key crops from the original image based on constraints generated using an image saliency map, 2. contrastive learning with a loss that forces models to look at the relevant object regions that are common between the query and key crops through Grad-CAM supervision. While our approach is generic and can be applied to any architecture, we describe \app{} in context of the Momentum Contrast (MoCo)~\cite{he2019moco} pretraining setup.

MoCo learns to perform the instance discrimination~\cite{Wu_2018_Instance} task using the InfoNCE~\cite{oord2018cpcv1}  contrastive objective (described in detail later in Section~\ref{sec:cast_loss}). In this task, a query and a key form a positive pair if they are data-augmented versions of the same image, and form a negative pair otherwise.  
MoCo builds a dynamic dictionary of negatives with a queue and a moving-averaged encoder which enabled access to a large and consistent dictionary which can be utilized for contrastive learning of representations.

\begin{figure}[t]
    \centering
    \includegraphics[width=\linewidth]{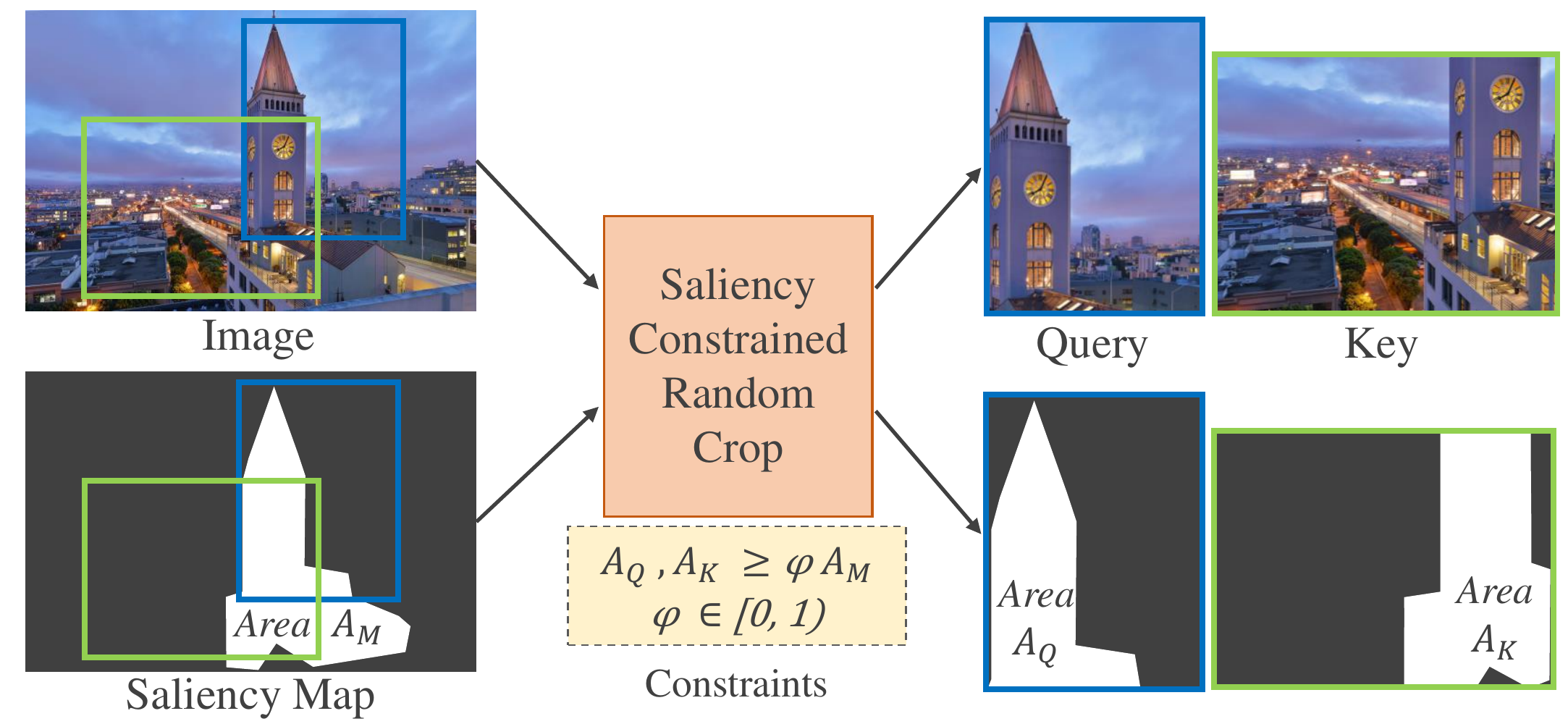}
    \vspace{-10pt}
    \caption{
        \textbf{Saliency-constrained Random Crop.} We compute query and key crops based on a saliency map and specified area constraints. } 
    \label{fig:saliency_crop}
    \vspace{-15pt}
\end{figure}

\subsection{Saliency-constrained Random Cropping}
\label{subsec:cropping}

We aim to improve the visual grounding ability of self-supervised models by explicitly supervising models to look at relevant image regions.
We provide this supervision as a \emph{saliency map}---a binary mask indicating these relevant image regions.
These typically contain all the objects and other important visual concepts present in the image. 
In this work we utilize Deep-USPS~\cite{nguyen2019deepusps} to generate unsupervised saliency maps. 
But providing localization supervision is not sufficient to fix visual grounding.
As shown in \Cref{fig:scene_level_images}, models often receive a noisy training signal---random crops from an image may contain different objects, or none at all.
To fix this problem, we design a random crop transform that generates input crops constrained to overlap with the saliency map.

\paragraph{Crop Constraints:}
Given an input image $I$ with height $h$ and width $w$, the standard data augmentation involves sampling two independent random crops (query and key) for input to the model.
Here, we assume access to a saliency map $M \in \{0, 1\}^{h \times w}$, where $M_{ij} = 1$ indicates pixel $(i, j)$ is salient, and area of salient region is $A_M = \sum_{i,j} M_{i,j}$.

Consider the example in \Cref{fig:saliency_crop}.
Our technique samples random crops based on a constraint specified by a hyperparameter $\phi \in [0, 1)$ :
\emph{the area of saliency map $M$ covered by each crop must be at least $\phi \cdot A_M$.}

We refer $\phi$ as the \emph{area-overlap threshold}.
Higher values of $\phi$ imply stricter constraints---enforcing higher overlap between sampled crops and salient regions,
whereas setting $\phi = 0.0$ recovers the unconstrained random crop, used by \moco{} and other existing SSL methods.

As seen in \Cref{fig:saliency_crop}, this simple area-overlap based constraint ensures that both the query and key crops contain some salient regions,
and we supervise models to focus on them during training to improve visual grounding.



\begin{figure*}[t]
    \centering
    \includegraphics[width=0.98\textwidth]{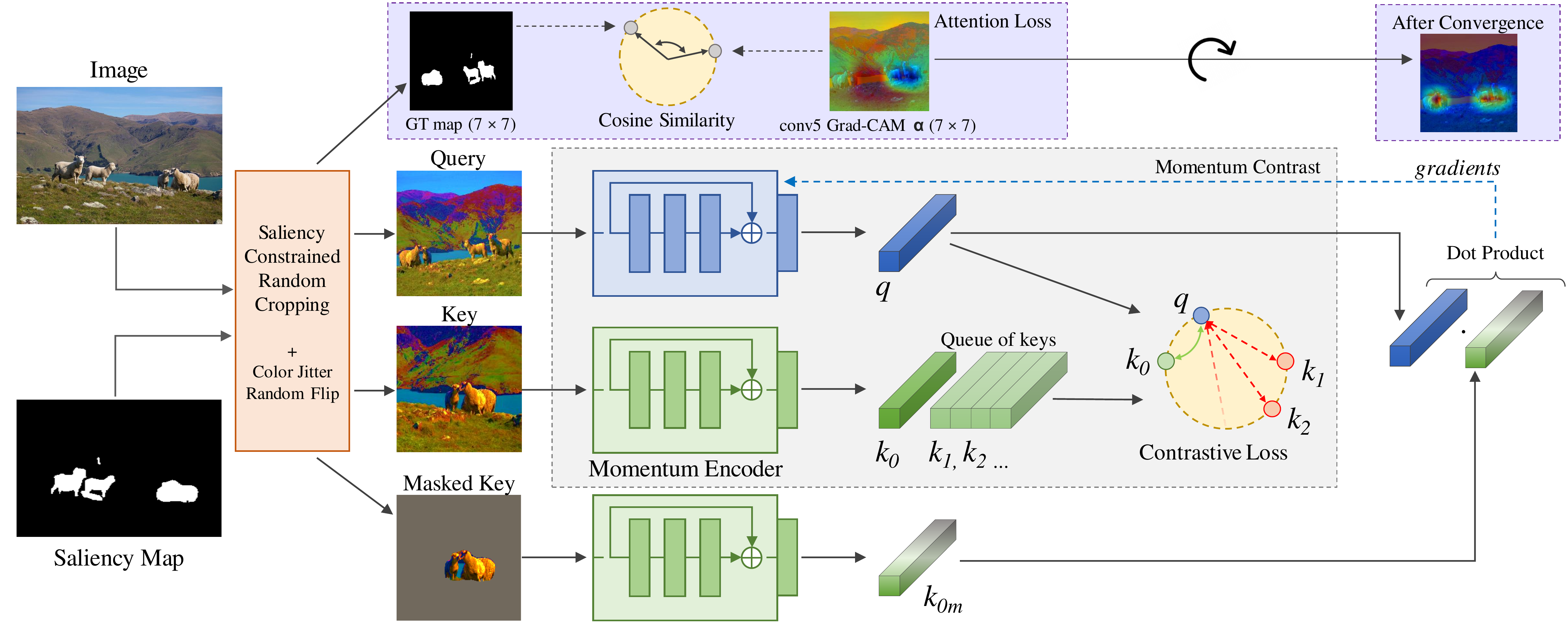}
    \vspace{-5pt}
    \caption{\textbf{Contrastive Attention-Supervised Tuning (CAST):} Given an image (top-left), we compute a saliency map (bottom-left), which we use to generate query and key crops and their corresponding saliency maps. We obtain the query and key feature representations with a forward-pass through the encoder network (blue box) and momentum encoder, respectively. Through the contrastive loss, we pull the representations of query and key crop together, while pushing the query representation away from other representations in a dynamic queue. We pass the salient regions in the key crop through the same momentum encoder, and compute the dot product between query and masked key representation. We then compute its gradient with respect to the last encoder convolution layer and weigh the forward activation maps to get Grad-CAM map. Finally, we add an attention loss that encourages Grad-CAM to look at all the salient image regions in the query crop.}
    \label{fig:pipeline}
    \vspace{-10pt}
\end{figure*}


The premise of our approach is that when contrastive models such as MoCo~\cite{he2019moco}  are given multiple crops from an image, focusing on the salient (object) regions in the crops would make them learn representations that are more generalizable. 
These models are likely to be more grounded, and are thus less likely to learn unwanted biases. 
\app{} introduces a grounding loss that encourages this behaviour. 

Recall that MoCo samples two crops, \textit{query} and \textit{key}, and enforces their representations to be closer compared to the other representations in a large dynamic queue. 
The random cropping transformations (shown in \reffig{fig:saliency_crop}) used to obtain the query and key crop can also be applied to the image-specific saliency map, $M$.
This results in two corresponding saliency maps $M_q$ and $M_k$, each containing the salient object regions in the \textit{query} and \textit{key} crop. 


However, the entirety of the object may not exist in both the \textit{query} and the \textit{key} crop. 
Hence, when considering the saliency map corresponding to the \textit{query}, $M_q$, there can be cases where only a part of the salient region in the query exists in the key. 
In such scenarios, we consider \textit{all} the regions in the query that correspond to the salient regions in key. See example in \reffig{fig:pipeline}, where the two crops contain varying extent of the sheep. 
In such cases, the saliency map corresponding to the query would contain all regions in the query that contain the sheep. 
Qualitative examples showing how our random crops differ from regular random crops used by self-supervised learning methods can be found in \reffig{fig:sup_random_crops} in the Appendix. 
As described next, we use these saliency maps to supervise where networks attend to.

\subsection{Computing Network Importance}
We define network importance as the importance placed by the encoder network, $f_q$ on the spatial regions of the query, $x^q$, in order to predict that the query representation is closest to the representation of the key $x^k$, as compared to all the representations present in the queue. To compute network importance, we extend Grad-CAM~\cite{selvaraju2017gradcam} to contrastively-trained models. 

To obtain Grad-CAM, we first forward-propagate the query crop $x^q$ to query encoder $f_q$ (see  blue encoder box in \reffig{fig:pipeline}).
The key is then masked with the corresponding saliency map to obtain the salient regions in the key crop (see the bottom part of \reffig{fig:pipeline}). 
This masked key, $x^k_m  = x^k * M_k$, is then fed to the key encoder (see green encoder box in \reffig{fig:pipeline}), $f_k$. 
Following MoCo, we dot-product the query representation, $q = f_q(x^q)$, with the masked key crop representation $k^{m}_{+} = f_k(x^k * M_k)$, and each of the other representations in the dynamic queue, and concatenate them. 
We then one-hot encode the dot-product for the correct key and compute its gradients (shown as blue backward arrow in \reffig{fig:pipeline}) \wrt last convolutional layer activations of the encoder network, $A_{conv5}^{f_q}$, as, 
\vspace{-2pt}
\begin{equation}\label{eq:alpha}
    \alpha{}_{q} =
    \overbrace{
        \sum_{i,j}
    }^{\text{global pooling}} \mkern-65mu
    \hspace{10pt}
    \underbrace{         
        \frac{\partial q \cdot k^{m}_{+}}{\partial A_{conv5}^{f_q}}
    }_{\text{gradients via backprop}}\vspace{-8pt}
\end{equation}

As in Grad-CAM~\cite{selvaraju2017gradcam}, the $\alpha_q$ values indicate the importance of each of the last convolutional layer neurons, $n$, in the encoder network for matching the query and masked key representation. To get the regions represented by these important convolutional neurons, we use  $\alpha_q$ to perform a weighted combination of forward activation maps corresponding to query, $A_{conv5}^{f_q}$, followed by a ReLU to obtain,
\vspace{-3pt}
\begin{equation} \label{eq:gcam}
    \text{ \cal G}_q = ReLU \underbrace{\left(\sum_{n} \alpha{}_{\text{q}} A_{conv5}^{f_{\text{q}}}\right)}_{\text{linear combination}}
\end{equation}

The higher values in the resulting Grad-CAM map (\reffig{fig:pipeline} top-right) indicates query regions which the network relies on when mapping the masked key regions, $x^k * M_k$, to the entire query crop, $x^q$. 
These heatmaps form the basis for enforcing attention supervision, which we explain next. 

\begin{table*}[h]
    \newcommand{\apbbox}[1]{AP$^\text{bbox}_\text{#1}$}
    \newcommand{\apmask}[1]{AP$^\text{mask}_\text{#1}$}
    \newcolumntype{Y}{>{\raggedright\arraybackslash}X}
    \newcolumntype{Z}{>{\centering\arraybackslash}X}

    \centering
    \footnotesize
    \setlength\tabcolsep{1pt}
    \renewcommand{\arraystretch}{1.2}

    \begin{tabularx}{\textwidth}{c l c l c l c YYY c YYYYYY}
    \toprule
    ~
    & \multicolumn{1}{l}{\bf \multirow[b]{2}{*}{Method}}
    & ~~~
    & \multicolumn{1}{c}{\bf \vocclf{} clf.}
    &~~& \multicolumn{1}{c}{\bf \inclf{} clf.}
    &~~& \multicolumn{3}{c}{\bf \voc{} Detection}
    &~~& \multicolumn{6}{c}{\bf COCO Instance Segmentation} \\
    \cmidrule{4-4} \cmidrule{6-6} \cmidrule{8-10} \cmidrule{12-17}

    ~ & ~ & ~ & \multicolumn{1}{c}{mAP} & ~ & \multicolumn{1}{c}{Top-1 acc.} & ~ &
                \apbbox{all} & \apbbox{50} & \apbbox{75} &&
                \apbbox{all} & \apbbox{50} & \apbbox{75} &
                \apmask{all} & \apmask{50} & \apmask{75} \\
    \midrule
    \band
    \ttbf{1)}  & \random{} &&
                -- && -- &&
                33.8 & 60.2 & 33.1 &&
                36.7 & 56.7 & 40.0 & 33.7 & 53.8 & 35.9 \\

    \ttbf{2)}  & \mocococo{} &&
                66.9 && 46.5 &&                             
                47.5 & 75.4 & 51.5 &&                       
                38.3 & 58.7 & 41.5 & 34.9 & 55.7  & 37.2 \\  
    
    \ttbf{3)}  & \hspace{5pt} + Constrained Crop &&
                69.3\Rise{2.3} && 46.0\Drop{0.5} &&                             
                49.0\Rise{1.5} & 77.4\Rise{2.0} & 52.4\Rise{0.9} &&                       
                38.3\rise{0.0} & 58.7\rise{0.0} & 41.6\rise{0.1} & 34.8\drop{0.1} & 55.7\rise{0.0} & 37.2\rise{0.0} \\  
    
    \ttbf{4)}  & \hspace{5pt} + \app{} &&
                73.1\Rise{6.2} && 48.7\Rise{2.1} &&                             
                54.2\Rise{6.7} & 80.1\Rise{4.7} & 59.9\Rise{8.4} &&                       
                39.4\Rise{1.1} & 60.0\Rise{1.3} & 42.8\Rise{1.3} & 35.8\Rise{0.9} & 57.1\Rise{1.4} & 38.6\Rise{1.4} \\  
    \bottomrule
    \end{tabularx}
    \vspace{-5pt}
    \caption{\textbf{Transfer Learning on Downstream Tasks:}
    We report results on four downstream tasks.
    For every task, all methods use the same architecture and learning setup.
    \vocclf{} and \inclf{} use frozen feature extractor, COCO Instance Segmentation and \voc{} Detection involve end-to-end fine-tuning.
    Gaps with \mocococo{} are shown on the side (differences $\ge 0.5$ are colored).
    We observe that training with \app{} outperforms all baselines by a huge margin on all downstream tasks.} 
    \label{tab:main_results}
    \vspace{-15pt}
\end{table*}

\subsection{CAST Loss}
\label{sec:cast_loss}
The CAST loss consists of two components: 1. a Contrastive loss, $L_{cont}$ from \cite{he2019moco}, that measures the similarities of original sample pairs ($x_q$ and $x_k$) in the representation space (yellow circle in \reffig{fig:pipeline}), and 2. an Attention loss, $L_{att}$, that measures the similarity of Grad-CAM heatmap to its corresponding saliency map, $M_q$ (purple box in \reffig{fig:pipeline}). $L_{cont}$ is defined as 
\vspace{-5pt}
\begin{equation} \label{eq:contrastive_loss}
    \cal L_{\text{cont}} = - \text{log} \frac{\text{exp}\left( \text{q} \cdot \text{k}_{\text{+}}/\tau\right)} {\sum_{\text{i=0}}^{K} \text{exp}\left(\text{q}\cdot \text{k}_{\text{i}}/\tau\right) } 
\end{equation}
where $K$ denotes the number of  representations in the queue and $\tau$ is the temperature hyperparameter. 

As network importances (from above) are gradient based, we penalize errors in the predicted Grad-CAM map, $G_q$ based on cosine distance---emphasizing alignment over magnitude (see the top box in \reffig{fig:pipeline}). We minimize the cosine distance loss as,
\vspace{-5pt}
\begin{equation}
{\cal L_{\text{att}}} =  1 - \frac{ G_{q} \cdot M_{\text{q}}} {\lVert G_{q} \rVert~\lVert M_{\text{q}}\rVert}
\end{equation}

The final \fullapp{} loss becomes $\mathcal{L_{CAST}} =  L_{cont} + \lambda L_{att}$
 
\noindent The second term encourages the network to base predictions on the correct regions and the first term encourages the network to actually make the right prediction. 
Note that $A^{f_q}_{conv5}$ is a function of all the encoder parameters until last convolution layer and $\alpha{}_{q}$ is a function of the layers from the last convolutional layer until the final fully-connected layer, and the key encoder features. 
They keys, $k$ and $k_m$, are detached from the key encoder, and therefore gradients do not get passed through them. 
Hence, while Grad-CAM is a function of of the both the query and the key encoder weights, during the update through an optimization algorithm, only the weights on the query encoder are updated. 
In MoCo, since the key encoder is a moving average of the query encoder, the key encoder weights get updated eventually during training.
 
\section{Evaluation}
\label{sec:downstream_eval}

In our experiments, we aim to show that training self-supervised models with localization supervision offers two benefits---better visual grounding ability, and better transfer learning performance.
We pretrain \moco{}~\cite{he2019moco} with \app{} on images from the COCO dataset~\cite{lin2014microsoft}, and then evaluate the
transfer performance and grounding ability of learned features on multiple downstream tasks.


\subsection{Transfer Learning on Downstream Tasks}
\label{section:downstream_eval}

First, we evaluate the quality of the learned features by transferring them to four downstream visual recognition tasks:
\textbf{(a)} \voc{}~\cite{everingham2009voc} linear classification, \textbf{(b)} \imagenet{}~\cite{deng2009imagenet,russakovsky2015imagenet} linear classification, \textbf{(c)} \voc{} object detection, \textbf{(d)} COCO~\cite{lin2014microsoft} instance segmentation.
Consistent with prior SSL research, our downstream tasks involve learning setups where the pretrained network is used as either a frozen feature extractor \textbf{(a, b)}, or weight initialization for fine-tuning \textbf{(c, d)}.

\noindent \textbf{Baselines:}
We compare \mocococo{} + \app{} with baseline methods to show the importance of different components of our algorithm:
\begin{compactenum}[\hspace{1pt}1.]
    \item \textbf{\random{}} uses no pretrained visual features.
    \item \textbf{\mocococo{}}, without \app{} attention loss ($\lambda = 0$) and constrained random cropping ($\phi = 0$).
    \item \textbf{\mocococo{} + Constrained Crop}, without \app{} attention loss, to observe gains from better cropping alone.
\end{compactenum}

\noindent {For all tasks, we follow the same hyperparameters as \virtex{}~\cite{desai2020virtex}, using its publicly available code~\footnote{Code available at: \url{https://github.com/kdexd/virtex}}.
\virtex{} uses a similar evaluation setup as the majority of recent work on self-supervised learning~\cite{goyal2019scaling,he2019moco,misra2019pirl,junnan2020pcl,caron2020swav}, including our primary baseline, \moco{}.
We describe the main details here.}

\noindent \textbf{\voc{} Linear Classification:}
We train on \vocclf{} \texttt{trainval} split and report mAP on \texttt{test} split.
We extract the $7 \times 7$ spatial grid of 2048-dimensional features from the last convolutional layer,
and downsample them to $2 \times 2$ grid via adaptive average pooling.
Then, we flatten and L2-normalize these features to yield 8192-dimensional features.
We train per-class SVMs for costs $C \in \{0.01, 0.1, 1.0, 10.0\}$, and select best $C$ by 3-fold cross validation.
Other SVM hyperparameters are same as \cite{desai2020virtex}.

\noindent \textbf{\imagenet{} Linear Classification}:
{We train on ILSVRC 2012 \texttt{train} split and report top-1 center crop accuracy on the \texttt{val} split.
We train a linear layer on 2048-dimensional global average pooled features extracted from the network.
We train for 100 epochs using SGD with momentum 0.9, weight decay 0, and with batch size 256 distributed across 8 Nvidia V100 GPUs.
Similar to \moco{}, we start with learning rate 30, and divide it by 10 at epochs 60 and 80.}

\noindent \textbf{\voc{} Object Detection:}
We train Faster R-CNN~\cite{ren2015faster} with ResNet-50-C4 backbone.
We initialize this backbone with pretrained weights, train on \texttt{trainval07+12} split, and evaluate on \texttt{test2007} split.
We train for 24K iterations using SGD with momentum 0.9, batch size 16 (2 per GPU), and weight decay $10^{-4}$.
We use maximum learning rate 0.02, perform linear warmup for first 100 iterations, and divide it by 10 at iterations 18K and 22K.
We fine-tune the network end-to-end, with batch normalization layers synchronized across GPUs (\emph{SyncBN})~\cite{peng2018megdet}.

\noindent \textbf{COCO Instance Segmentation:}
We train Mask R-CNN~\cite{he2017mask} models with ResNet-50-FPN backbones~\cite{lin2017feature} on \texttt{train2017} split, and evaluate on \texttt{val2017} split.
We follow $2\times$ training schedule implemented in Detectron2~\cite{wu2019detectron2}, and fine-tune with SyncBN in the backbone and FPN layers.

\noindent \textbf{Results:} We summarize our results in \Cref{tab:main_results}. MoCo + CAST outperforms MoCo on all downstream tasks, obtaining robust gains on classification, detection, and instance segmentation. The performance improvement is especially large on the VOC detection task, aided by the improved visual grounding in models trained with CAST.  We also find that our unsupervised saliency-constrained cropping alone outperforms MoCo on \vocclf{} and VOC-Detection, and gets close to MoCo performance on Imagenet-1k and COCO instance segmentation tasks.  

\begin{table}[t]
    \centering \footnotesize
    \setlength{\tabcolsep}{4pt}

    \begin{subfigure}[t]{\linewidth}
        \begin{tabularx}{\linewidth}{lXcc}
        \toprule
        \textbf{Area threshold} && $\phi= 0.2$ & $\phi = 0.0$ \\
        \midrule
        \textbf{\vocclf{}} && \graycell 73.1 & 72.3\Drop{0.8} \\
        \textbf{\inclf{}} && \graycell 48.7 & 47.4\Drop{1.3} \\
        \bottomrule
        \end{tabularx}
        \vspace{-3pt}
        \caption{Effect of Area threshold $\phi$ (Fixing $\lambda = 3.0$)}
        \label{subtab:ablations1}
        \vspace{5pt}
    \end{subfigure}

    \begin{subfigure}[t]{\linewidth}
        \begin{tabularx}{\linewidth}{lXcccc}
        \toprule
        \textbf{Loss weighing factor} && $\lambda = 0.0$ & $\lambda = 1.0$ & $\lambda = 3.0$ & $\lambda = 5.0$ \\
        \midrule
        \textbf{\vocclf{}} && \graycell 66.9 & 72.6\Rise{5.7} & 73.1\Rise{6.2} & 72.4\Rise{5.5} \\
        \textbf{\inclf{}} && \graycell 46.5 & 48.7\Rise{2.2} & 48.7\Rise{2.2} & 47.6\Rise{1.1} \\
        \bottomrule
        \end{tabularx}
        \vspace{-3pt}
        \caption{Effect of loss weighing factor $\lambda$ (Fixing $\phi = 0.2$)}
        \label{subtab:ablations2}
        \vspace{5pt}
    \end{subfigure}

    \begin{subfigure}[t]{\linewidth}
        \begin{tabularx}{\linewidth}{lXcc}
        \toprule
        \textbf{\moco{} Projection Layer} && 1-layer Linear & 2-layer MLP \\
        \midrule
        \textbf{\vocclf{}} && \graycell 73.1 & 72.8\drop{0.3} \\
        \textbf{\inclf{}} && \graycell 48.7 & 50.1\Rise{1.4} \\
        \bottomrule
        \end{tabularx}
        \vspace{-3pt}
        \caption{Effect of improving underlying \mocococo{}}
        \label{subtab:ablations3}
        \vspace{5pt}
    \end{subfigure}

    \begin{subfigure}[t]{\linewidth}
        \newcommand{\apbbox}[1]{AP$^\text{bbox}_\text{#1}$}
        \newcommand{\apmask}[1]{AP$^\text{mask}_\text{#1}$}
        \newcolumntype{Y}{>{\raggedright\arraybackslash}X}
        \newcolumntype{Z}{>{\centering\arraybackslash}X}
    
        \centering
        \footnotesize
        \setlength\tabcolsep{1pt}
        \renewcommand{\arraystretch}{1.1}
    
        \begin{tabularx}{\textwidth}{l c X c X c YYY}
        \toprule
        \multicolumn{1}{l}{\bf \multirow[b]{2}{*}{Supervision}}
        &~~~& \multicolumn{1}{c}{\bf \vocclf{}}
        &~~~& \multicolumn{1}{c}{\bf \inclf{}}
        &~~~& \multicolumn{3}{c}{\bf \voc{} Detection}\\
        \cmidrule{3-3} \cmidrule{5-5} \cmidrule{7-9}
    
        ~ & ~ & mAP & ~ & Top-1 & ~ & \apbbox{all} & \apbbox{50} & \apbbox{75}\\
        \midrule
        \band Query     && 73.1 && 48.7 &&                             
                            54.2 & 80.1 & 59.9 \\
        Intersection    && 72.0\Drop{1.1} && 49.4~\Rise{0.7} &&         
                            53.3~\Drop{0.9} & 79.7~\drop{0.4} & 59.0~\Drop{0.9} \\
        \bottomrule
        \end{tabularx}
        \vspace{-3pt}
        \caption{Effect of suppressing saliency supervision}
        \label{subtab:ablations4}
        \vspace{5pt}
    \end{subfigure}
    \vspace{-5pt}
    \caption{
        \textbf{Ablations for \mocococo{}  + \app{} training:}
        We conduct ablation studies to isolate the effects of our training components.
        \textbf{(a)} Replacing saliency-constrained random cropping with default version from \moco{} ($\phi = 0.0$) hurts performance.
        \textbf{(b)} Increasing weight of \app{} loss generally improves performance up to a point ($\lambda = 1.0, 3.0$).
        \textbf{(c)} Adding known improvements to underlying \moco{} model (MLP layer) also transfer to \app{}.
        \textbf{(d)} Restricting attention supervision to only the intersection of query and key hurts downstream performance.
    }
    \label{tab:train_ablations}
    \vspace{-10pt}
\end{table}



\begin{figure}[h]
    \centering
    \includegraphics[width=0.47\textwidth]{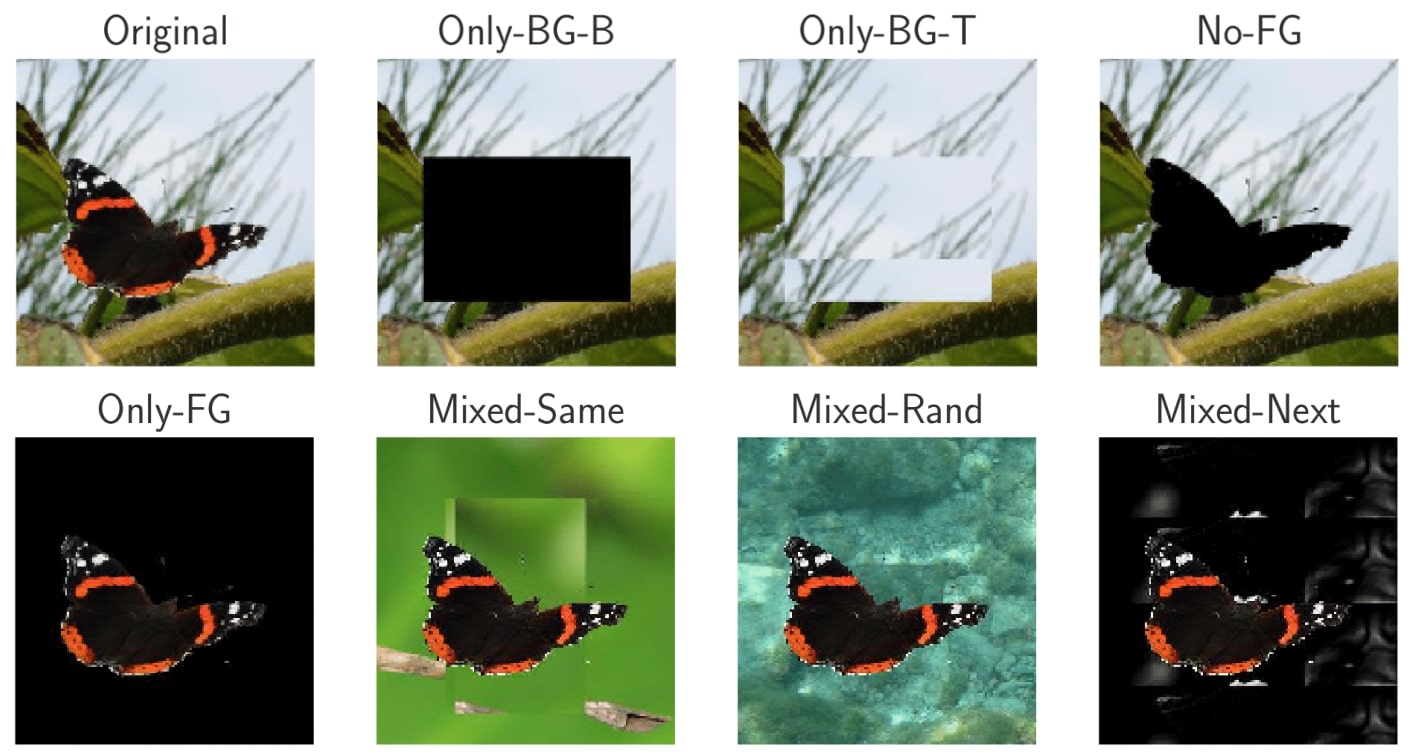}
    \caption{We evaluate CAST using the Backgrounds Challenge~\cite{xiao2020noise} dataset designed to evaluate background-robustness of models. FG = Foreground, BG = background. Foreground-background combinations include: Only-BG-B (FG:  Black, BG: Unmodified), Only-BG-T (FG: Tiled background, BG: Unmodified), Mixed-Same (FG: Unmodified,  BG: Random BG of the same class), Mixed-Rand (FG: Unmodified, BG: Random BG of a random class), and Mixed-Next (FG: Unmodified, BG: Random BG of the next class.) }
    \label{fig:backgrounds}
    \vspace{-14pt}
\end{figure}

\begin{table*}[t]

    \newcolumntype{Y}{>{\raggedright\arraybackslash}X}
    \newcolumntype{Z}{>{\centering\arraybackslash}X}

    \centering
    \footnotesize
    \renewcommand{\arraystretch}{1.2}

    \begin{tabularx}{\textwidth}{ p{0.15\textwidth} c llllllll}
    \toprule
     \multicolumn{1}{l}{\bf \multirow[b]{2}{*}{\makecell{MoCo-COCO\\ Performance}}}
    & 
    & \multicolumn{8}{c}{\bf Backgrounds Challenge Setting}\\
    \cmidrule{3-10} 
     ~ &  ~& Original & \makecell{Mixed-Same}  & \makecell{Mixed-Rand}  & \makecell{Mixed-Next} &   \makecell{Only-FG}  &
     \makecell{No-FG} & \makecell{Only-BG-B}  &  \makecell{Only-BG-T}    \\
    \midrule
    \band
    Default &~& 72.62 & 45.75 & 30.44 & 26.86 & 30.42 & 23.95 &  5.06 & 12.62  \\
    + Constrained-Crop   &~&      74.79\Rise{2.17}   &  52.64\Rise{6.89}     & 39.14~\Rise{8.70}    & 34.17\Rise{7.31}    &  33.73\Rise{3.31}   & 22.74\Riseneg{1.21}     &  4.10\Riseneg{0.96}    &  11.88\Riseneg{0.74}  \\
    + CAST  &~&      77.33\Rise{4.71}   &  54.42\Rise{8.67}     & 39.93\Rise{9.49}      & 37.46\Rise{10.60}  &  43.26\Rise{12.84}     & 23.70\Riseneg{0.15}    &  4.40\Riseneg{0.66}    &  12.59\riseneg{0.03}   \\
    \bottomrule
    \end{tabularx}
    \caption{\app{} obtains large improvements over MoCo on the Backgrounds Challenge, a 9-class image classification dataset  containing foreground objects  superimposed  on  various  background  types. In settings where the foreground is present (columns 1-5), CAST's visual grounding ability leads to substantial performance gains. When foreground is absent (columns 6-8), CAST performs slightly worse, validating that CAST-trained models learn fewer background correlations.}
    \label{tab:backgrounds_results}
\end{table*}

\begin{figure*}[t]
\vspace{-10pt}
    \begin{subfigure}[t]{\textwidth}
    \includegraphics[width=\textwidth]{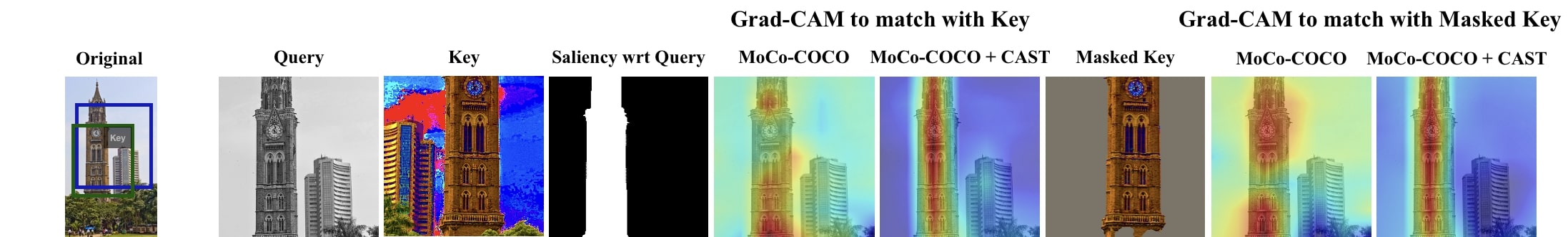}
    \vspace{1pt}
    \end{subfigure}
    \begin{subfigure}[t]{\textwidth}
    \includegraphics[width=\textwidth]{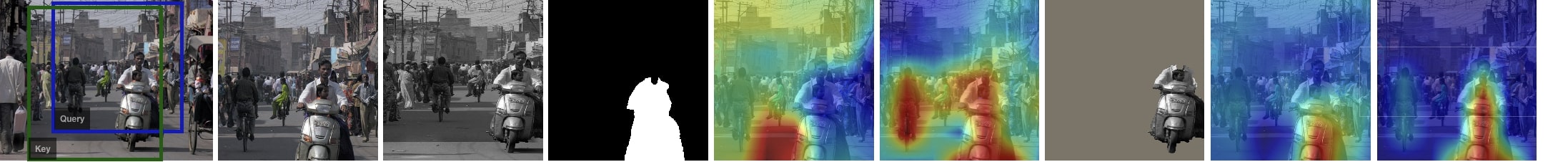}
    \vspace{1pt}
    \end{subfigure}
    \begin{subfigure}[t]{\textwidth}
    \includegraphics[width=\textwidth]{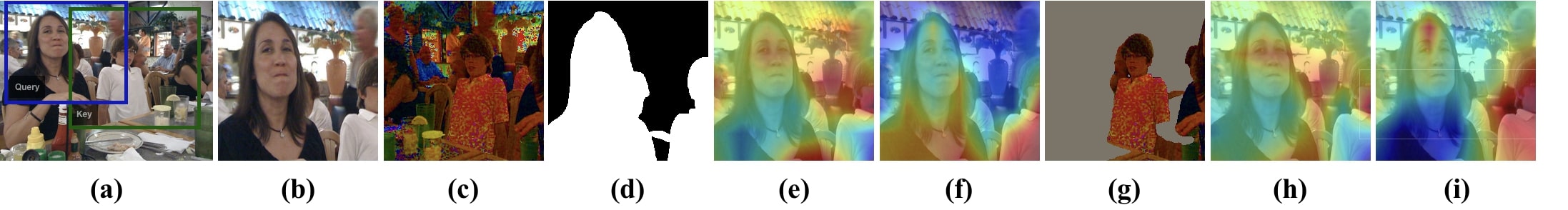}
    \vspace{1pt}
    \end{subfigure}
    \vspace{-25pt}
    \caption{CAST improves visual grounding of the contrastive self-supervised feature encoder. 
    Column (d) shows the saliency map according to the query crop (b). Grad-CAM visualizations in columns (e, f) show the query regions that MoCo and MoCo + CAST models rely on, in order to match the key crop (c). Finally, the MoCo and MoCo + CAST models rely on query regions (h) and (i) to match with the masked key representation (g). The example in top row shows that MoCo also looks at the sky in the background to match the masked key to the tower image in the query, indicating it has learnt spurious correlations. In contrast, the MoCo + CAST model looks just at the salient tower region. Similar trends are seen in the second row. In the third row, where there are two women in the foreground, the two crops contain different women. The MoCo + CAST model is able to localize the woman in black when matching the woman in white (see masked key), indicating that it has learned semantic category-specific representations. The baseline MoCo model looks primarily at the background regions.}
    \label{fig:pretrained_grounding}
    \vspace{-10pt}
\end{figure*}

\subsection{Ablation Studies}

Next, we conduct ablation studies on our training setup to isolate the effect of our design decisions.
In all these comparisons, we treat \mocococo{} with \app{} trained with default hyperparameters as our base model.
We mainly observe downstream performance of all ablations on \vocclf{} and \inclf{} linear classification setups.

\noindent \textbf{Effect of area threshold $\phi$:}
We use area-overlap based constraints conditioned on saliency maps for sampling random crops, specifying them via an area threshold hyperparameter $\phi$.
Here, we quantify the downstream performance improvement due to \emph{better} training supervision from strategically sampled crops---we train a model with $\phi = 0.0$ to recover the default random crop used in \moco{}.
Results are in \Cref{subtab:ablations1}, we observe that removing saliency-constrained random cropping hurts performance, indicating that our saliency-constrained random cropping technique indeed provides better training signal.

\noindent \textbf{Effect of loss weighing factor $\lambda$:}
As described in \Cref{sec:cast_loss}, the \app{} loss is a linear combination of contrastive and attention losses.
We combine them through a weighted sum, and use $\lambda$ to scale the attention loss.
Here, we experiment with different values of $\lambda$ with $\lambda \in \{0.0, 1.0, 3.0, 5.0\}$.
Note that $\lambda = 0.0$ means \mocococo{} + Constrained Crop (\Cref{tab:main_results}, row 2). Results from \Cref{subtab:ablations2}
show that non-zero values of $\lambda$ outperform $\lambda = 0.0$, indicating that attention loss is important in \app{}.
Higher $\lambda$ improve performance up to a point---performance improves with $\lambda = 1.0,3.0$, and slightly degrades with $\lambda = 5.0$.

\noindent \textbf{Effect of improving underlying \mocococo{}:}
\app{} is a general purpose method that can be added to contrastive SSL methods to improve their visual grounding.
Here, we investigate whether improving the underlying SSL method also shows improvements when trained with \app{}.
We consider \moco-v2 variant~\cite{chen2020improved}, replacing the linear projection with an MLP, inspired by SimCLR~\cite{chen2020simclr}.
Results from \Cref{subtab:ablations3} show that \moco-MLP + \app{} matches or exceeds \mocococo{} + \app{} on downstream tasks,
indicating that \app{} can provide additive improvements over its underlying SSL method.

\noindent \textbf{Effect of suppressing saliency supervision:}
We believe that focusing on salient image regions is important to improve visual grounding.
Hence, we force the model to focus on \emph{all} the salient regions inside query crop.
In contrast to our proposed approach, we train \mocococo{} + \app{} with \emph{reduced supervision} in this ablation study, enforcing the model to only look at salient regions inside the intersection of query and key crops. Results from \Cref{subtab:ablations4}
show that excluding some salient regions from the query crop (lying outside the intersection)
significantly hurts downstream performance on multiple tasks.
This indicates that looking beyond the common visual content between two crops to solve instance discrimination  yields better visual features.

\begin{figure*}[t]
    \begin{subfigure}[t]{0.5\textwidth}
    ~~~\includegraphics[width=0.95\textwidth]{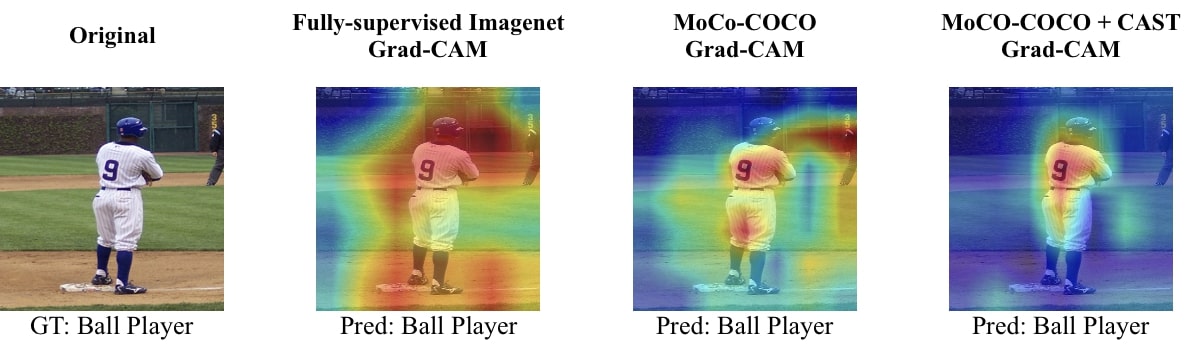}
    \vspace{-10pt}
    \caption{}
    \end{subfigure}%
    \begin{subfigure}[t]{0.5\textwidth}
    ~~~\includegraphics[width=0.95\textwidth]{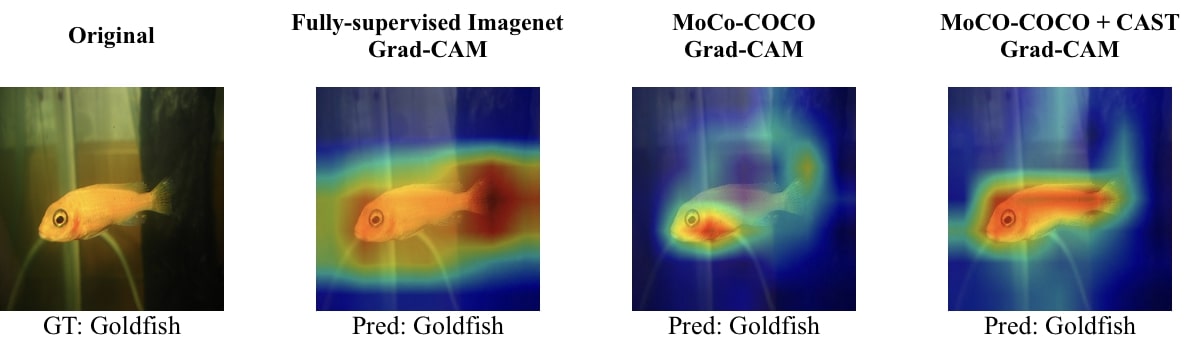}
    \vspace{-10pt}
    \caption{}
    \end{subfigure}
    \begin{subfigure}[t]{0.5\textwidth}
    ~~~\includegraphics[width=0.95\textwidth]{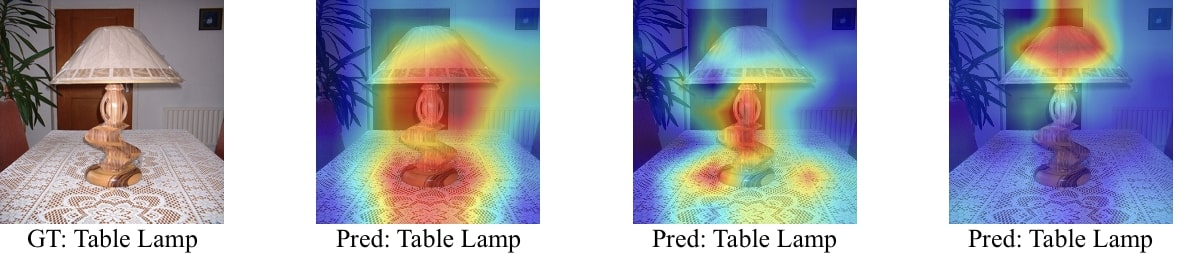}
    \vspace{-10pt}
    \caption{}
    \end{subfigure}%
    \begin{subfigure}[t]{0.5\textwidth}
    ~~~\includegraphics[width=0.95\textwidth]{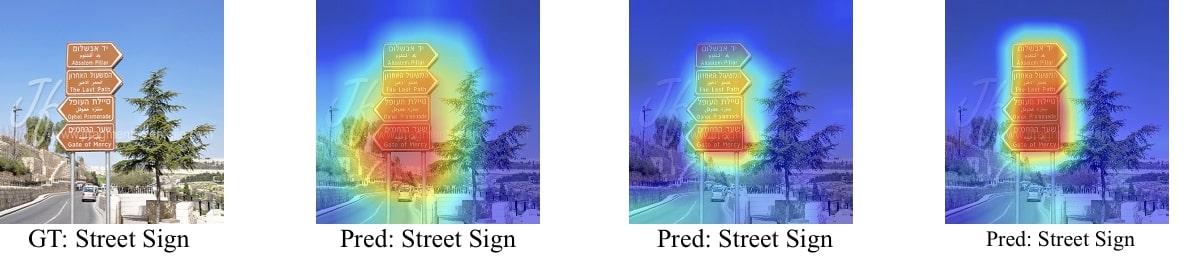}
    \vspace{-10pt}
    \caption{}
    \end{subfigure}
    \vspace{-10pt}
    \caption{Training with CAST also leads to improvements in grounding in downstream tasks. In this comparison of Grad-CAM attention maps from a fully supervised network and the self-supervised networks (MoCo and MoCo + CAST) on Imagenet-1k, we find that MoCo + CAST models tend to rely less on spurious correlations. (a) The MoCo + CAST model looks just at the player, whereas both fully supervised model and MoCo rely on the regions corresponding to the baseball field. (c) The MoCo  + CAST model looks only at the lamp, while other models also rely on the table below. (b, d) The MoCo + CAST model is much more precise at attending to the whole extent of the object of interest  as compared to other methods.}
    \label{fig:downstream_grounding}
    \vspace{-12pt}
\end{figure*}

\begin{figure}[h]
\vspace{-12pt}
    \centering
    \includegraphics[width=0.45\textwidth]{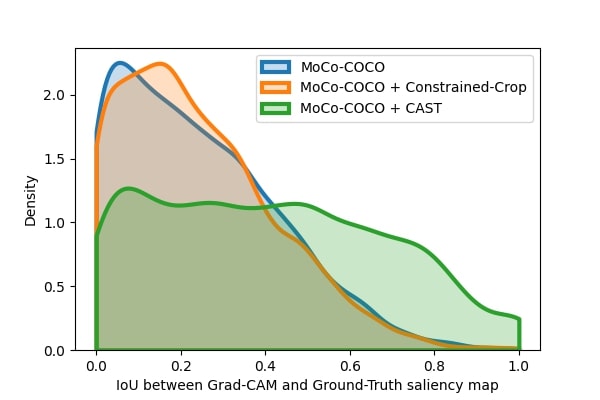}
    \vspace{-5pt}
    \caption{ CAST shows quantitative improvement in grounding for contrastive  self-supervised  models.  The distinct rightward shift in the green curve corresponding to the MoCo + CAST model shows that the gradient-based localization supervision loss significantly improves grounding.}
    \label{fig:visual_grounding_plot}
    \vspace{-12pt}
\end{figure}

\subsection{Evaluation on Backgrounds Challenge Dataset} \label{sec:backgrounds_challenge}
The Backgrounds Challenge~\cite{xiao2020noise} aims to assess the background-robustness of image classification models by measuring their accuracy on images containing foreground objects superimposed on various background types (see \cite{xiao2020noise} for details on dataset construction). The dataset consists of 9 ImageNet classes with 450 test images per class. The evaluations are performed on eight foreground-background combinations summarized in Figure~\ref{fig:backgrounds}. Since CAST forces a model to attend to salient objects during training, we expect a CAST-trained model to be less dependent on background correlations for classification.

We evaluate the performance of COCO-pretrained models on the Backgrounds Challenge using a linear layer trained with ImageNet-1K (as described in Section~\ref{section:downstream_eval}) using three settings: 1. MoCo,  2. MoCo trained with saliency-constrained random cropping alone, 3. MoCo trained with CAST (Table~\ref{tab:backgrounds_results}). Models trained with cropping constrains and with CAST outperform vanilla MoCo on all eight settings of the Backgrounds Challenge, with CAST obtaining the best performance on the five  settings where foreground is present. In the Only-FG setting, where background is set to black, CAST obtains an absolute improvement of 13\% over MoCo, indicating that CAST is significantly better at utilizing the foreground information, due to the saliency-driven attention-supervised training. In settings where background is swapped (Mixed-Same, Mixed-Rand, and Mixed-Next), CAST obtains 5-10\% absolute improvements, indicating that models trained with CAST are less dependent on background correlations. Finally, in settings that do not contain foreground objects (No-FG, Only-BG-B, and Only-BG-T), CAST performs slightly worse than the original model, as we would expect from a model that has learnt to rely less on the background signal in making classification decisions. 
Qualitative examples showing how CAST pretraining makes downstream models rely less on spurious background correlations can be found in Appendix \reffig{fig:sup_backgrounds_challenge}. 


\subsection{Evaluating Visual Grounding}
We use Grad-CAM for qualitative and quantitative evaluation of the visual grounding ability of a contrastive SSL model trained with CAST and its effect on grounding in downstream tasks. 
Examples in \reffig{fig:pretrained_grounding} show that the CAST-trained model seems to learn semantic category-specific feature representations, which allows it to look at objects of interest while performing query-key matching, and avoid learning spurious correlations. We quantify the improvement in grounding due to CAST using the COCO val split. First, we binarize the Grad-CAM maps by thresholding at 0.5. 
We then compute the intersection over union (IoU) between the Grad-CAM map and the saliency map corresponding to the query image. 
\reffig{fig:visual_grounding_plot} shows the density of IoU values for the baseline MoCo-COCO, MoCO-COCO with constrained cropping and MoCO-COCO with CAST. 
The mean IoU of the MoCo model trained with CAST over the COCO val set is 0.41, substantially larger than the mean IoU of the model trained  without CAST, which is 0.24. 
Moreover, the improvement in grounding ability is largely driven by the gradient-based localization supervision loss, as the mean IoU of a model trained with saliency-driven cropping constraints alone is also 0.24. 

Examples in ~\reffig{fig:downstream_grounding} shows how the improved grounding during pre-training translates to improved grounding in the downstream task of Imagenet linear classification. As seen in \reffig{fig:downstream_grounding} (a,c), the MoCo-COCO+CAST model relies less on spurious background correlations --- relying mostly on the player to predict Ball Player and the lamp to predict Table Lamp. \reffig{fig:downstream_grounding} (b,d) show that models pretrained with CAST learn to look at the whole extent of the object of interest as compared to other methods. 
More examples can be found in appendix (\reffig{fig:sup_pretrained_grounding} and \reffig{fig:supp_imagenet_grounding}).



\reducedSection{Conclusion}
We introduced a method for visually grounding contrastive self-supervised learning models, which improves feature representations learnt from scene images.
These feature representations are also less reliant on background correlations as compared to those trained with contrastive learning alone, which can lead to better out-of-distribution performance and greater robustness to contextual bias and adversarial backgrounds. We hope that our method leads to development of more general-purpose and robust self-supervised methods that  learn from noisy, unconstrained, real-world image data from the web.


\renewcommand{\thesection}{S\arabic{section}}
\renewcommand{\thetable}{S\arabic{table}}
\renewcommand{\thefigure}{S\arabic{figure}}

\pagebreak
\begin{appendices}

\reducedSection{Introduction}

This supplementary material is organized as follows. 
We first provide pseudo-code for our approach, Contrastive Attention-Supervised Tuning (CAST), showing how it can be easily applied on top of any contrastive self-supervised approaches. 
We then provide a sample of the kinds of random crops that are generated by our saliency-constrained random cropping approach and compare them to regular random crops used by  self-supervised learning methods. 
We then provide randomly sampled qualitative examples that show that CAST-trained feature representations  obtain better grounding, both on their own  and when finetuned on downstream tasks, such as image classification. 
We then show qualitative examples from the Background Challenge evaluation reported in Section 4.3 of the main paper. 


\begin{algorithm*}[t]
\newcommand{\hl}{\makebox[0pt][l]{\color{YellowOrange!15}\rule[-4pt]{0.6\linewidth}{9pt}}}

\caption{Pseudocode of \app{} applied on \moco{}, PyTorch-style. \app{} modifications highlighted in orange.}
\label{alg:code}
\definecolor{codeblue}{rgb}{0.25,0.5,0.5}
\lstset{
  backgroundcolor=\color{white},
  basicstyle=\fontsize{7.2pt}{7.2pt}\ttfamily\selectfont,
  columns=fullflexible,
  breaklines=true,
  captionpos=b,
  commentstyle=\fontsize{7.2pt}{7.2pt}\color{codeblue},
  keywordstyle=\fontsize{7.2pt}{7.2pt},
  escapeinside={<@}{@>},
}
\begin{lstlisting}[language=python]
# f_query, f_key: Encoder networks for query and key
# queue: Dictionary as a queue of K keys (CxK)
# m: momentum
# t: temperature

# Initialize key network (momentum encoder) parameters as query network.
f_key.params = f_query.params

# Load a mini-batch of images and saliency maps of N instances.
for (x, s) in dataloader:
    # Generate saliency-constrained random crops (+ other data augmentations).
    <@\hl@>x_q, s_q = aug(x, s)    # For MoCo: x_q = aug(x)
    <@\hl@>x_k, s_k = aug(x, s)    # For MoCo: x_k = aug(x)

    # --------------------------------------------------------------------------------
    # Compute MoCo contrastive loss:
    # --------------------------------------------------------------------------------

    # Query and key feature vectors.
    q = f_query.forward(x_q)  # . . . . . . . . . . . . . shape: (N, C)
    k = f_key.forward(x_k)  # . . . . . . . . . . . . . . shape: (N, C)
    k = k.detach()  # . . . . . . . . . . . . . . . . . . No gradient to keys.

    # Positive and negative logits.
    logits_pos = bmm(q.view(N, 1, C), k.view(N, C, 1))  # shape: (N, 1)
    logits_neg = bmm(q.view(N, C), queue.view(C, K))  # . shape: (N, K)

    logits = cat[logits_pos, logits_neg], dim=1)  # . . . shape: (N, 1+K)

    # MoCo contrastive loss (Positive labels at index 0).
    labels = zeros(N)
    loss = CrossEntropyLoss(logits/t, labels)

    # --------------------------------------------------------------------------------
    # Compute CAST loss:
    # --------------------------------------------------------------------------------

    # Generate masked key and forward through momentum encoder.
    <@\hl@>x_km = mask_op(x_k, s_k)
    <@\hl@>km = f_key.forward(x_km)  # . . . . . . . . . . . . . shape: (N, C)
    <@\hl@>km = km.detach()  # . . . . . . . . . . . . . . . . . No gradient to masked keys.

    # Compute dot product between query and masked key.
    <@\hl@>dot_prod = mm(q.view(N, 1, C), km.view(N, C, 1))  # . shape: (N, 1)

    # Get activations from conv5 layer of query network.
    <@\hl@>conv5_acts = lookup(f_query.params)  #. . . . . . . . shape: (N, 7, 7, 2048)

    # Compute gradients of these conv5 activations wrt this dot product.
    <@\hl@>conv5_grads = autograd.grad(dot_prod, conv5_acts)  #. shape: (N, 7, 7, 2048)

    # Compute Grad-CAM map from these gradients.
    <@\hl@>gradcam_map = compute_gradcam(conv5_grads)  # . . . . shape: (N, 7, 7)

    # CAST Loss: conv5 Grad-CAM and query saliency map (maximize similarity).
    <@\hl@>cast_loss = CosineSimilarityLoss(gradcam_map, resize7x7(s_q))
    <@\hl@>loss += lambda * cast_loss
    loss.backward()

    # SGD update for query network.
    update(f_query.params)

    # Momentum update for key network.
    f_key.params = m * f_key.params + (1 - m) * f_query.params

    # Update dictionary (do not add masked keys).
    enqueue(queue, k)  # Enqueue the current minibatch
    dequeue(queue)  # Dequeue the earliest minibatch
\end{lstlisting}
\end{algorithm*}

\section{Random Crops} \label{sup_random_crops}

\begin{figure*}[h!]
 \centering
  \begin{subfigure}{.5\textwidth}
  \centering
  \includegraphics[scale=0.25]{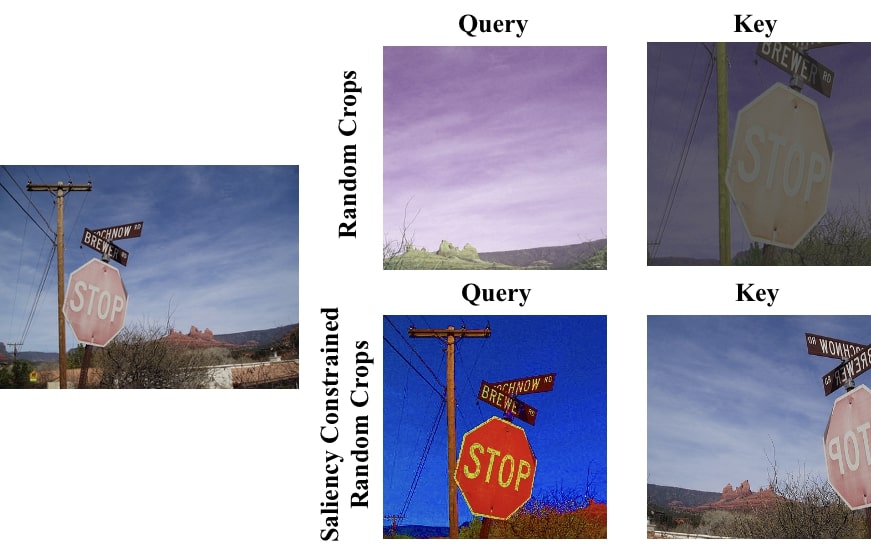}
  \caption{}
\end{subfigure}%
\begin{subfigure}{.5\textwidth}
  \centering
  \includegraphics[scale=0.25]{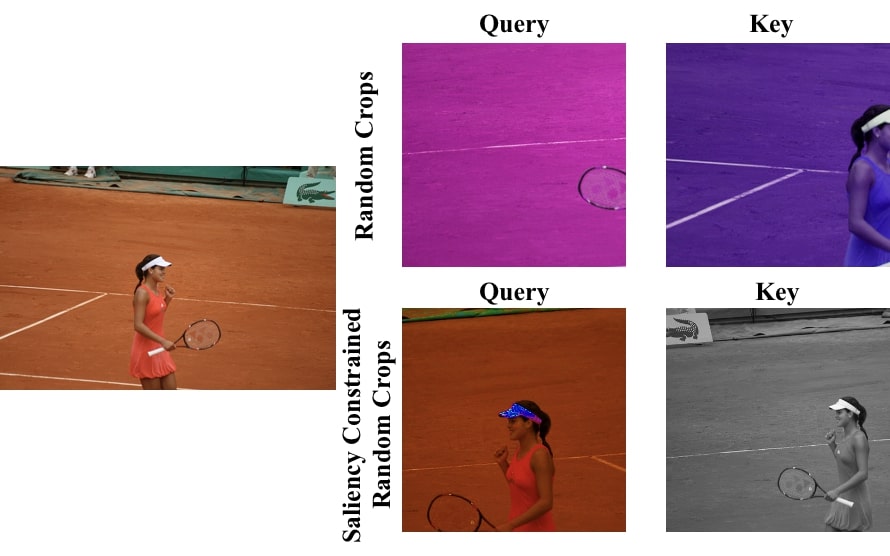}
  \caption{}
\end{subfigure}
 \begin{subfigure}{.5\textwidth}
  \centering
  \includegraphics[scale=0.25]{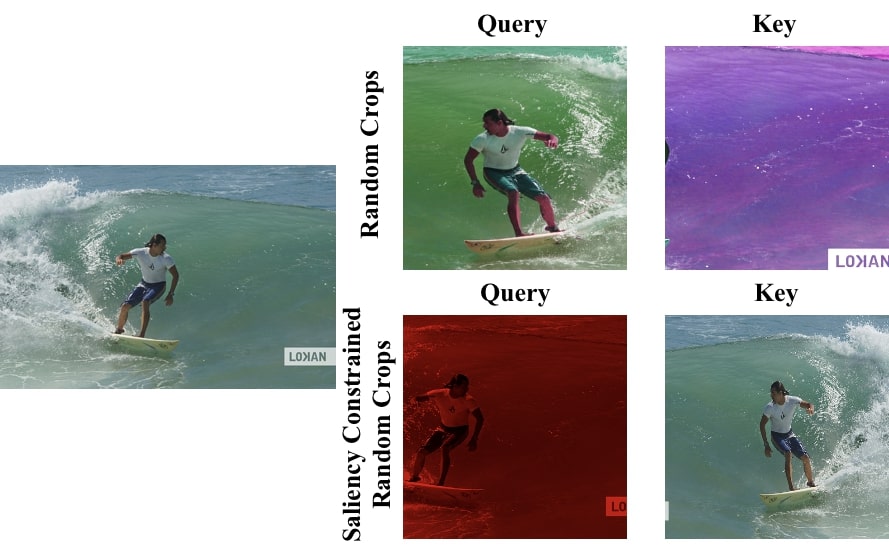}
  \caption{}
\end{subfigure}%
\begin{subfigure}{.5\textwidth}
  \centering
  \includegraphics[scale=0.25]{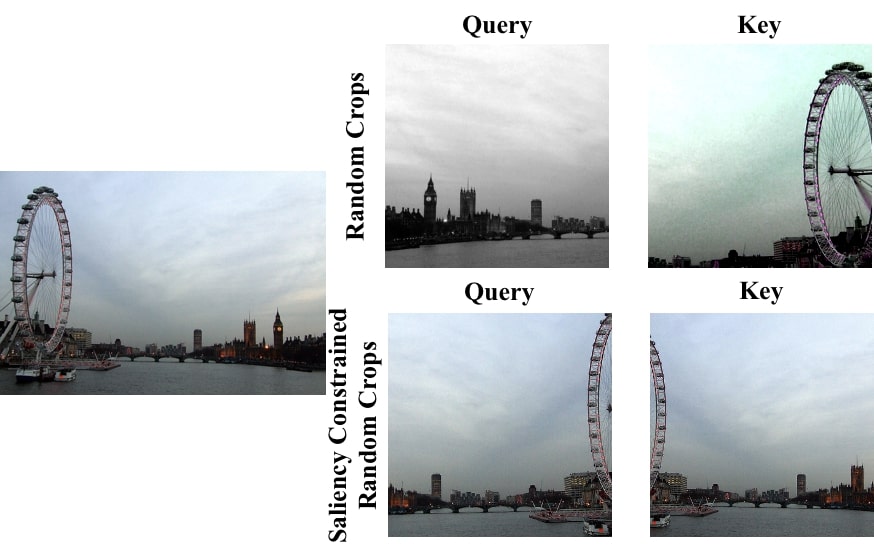}
  \caption{}
\end{subfigure}
\caption{Saliency-constrained random crops always have salient regions that are overlapping between two randomly sampled crops. This can be compared to unconstrained random crops used in MoCo. Notice how in the random crops of (a), the stop sign (salient region) is missing from query, however query and key crops from our saliency-constrained random cropping approach contain a part of the stop sign. Similarly in (b) the salient region corresponding woman is missing from one of the crops, and in (c), the salient regions corresponding to the surfer is not present in both the query and key crops. The crops which do not contain overlapping salient regions (top-row in each subfigure) tend to provide noisy training signal to contrastive approaches forcing them to incorrectly produce similar features for crops containing varying context, e.g.,  stop sign and sky in (a), surfer and waves in (c), etc. Saliency-constrained random cropping mitigates this problem.  
}
\label{fig:sup_random_crops}
\end{figure*}

As shown in \reffig{fig:scene_level_images} of the main paper, when training with complex scene-level images, models often receive a noisy training signal---random crops from an image may contain different objects, or none at all.
To fix this problem, we design a random crop transform that generates input crops constrained to overlap with the saliency map. 
In Figure~\ref{fig:sup_random_crops}, we show how our Saliency-Constrained Random Crops differ from unconstrained random crops used in MoCo and other contrastive approaches. 

\subsection{SSL Grounding}

\begin{figure*}[t]
    \begin{subfigure}[t]{\textwidth}
    \includegraphics[width=\textwidth]{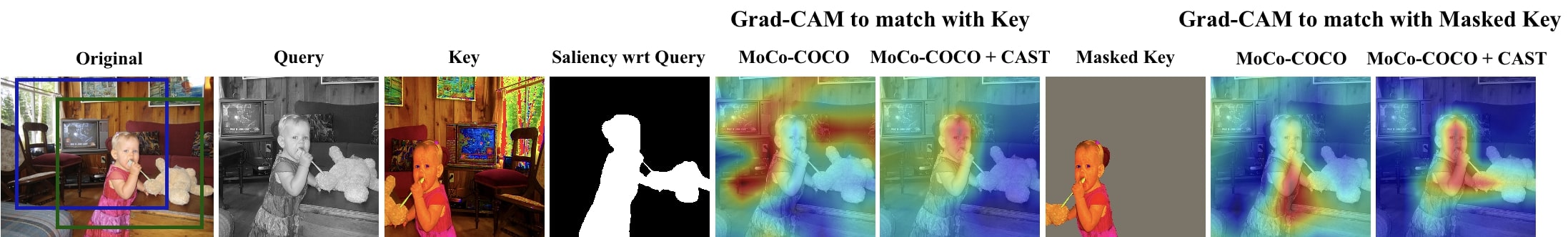}
    \vspace{4pt}
    \end{subfigure}
    \begin{subfigure}[t]{\textwidth}
    \includegraphics[width=\textwidth]{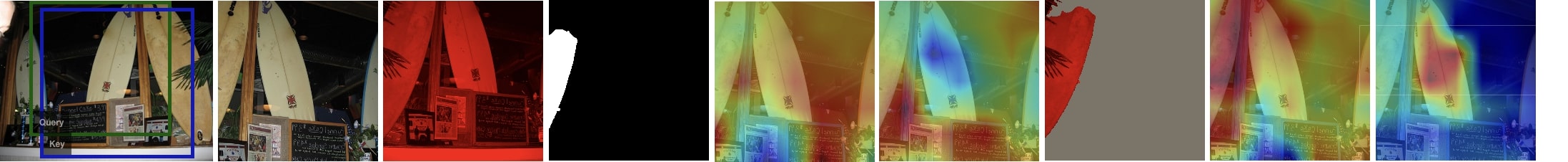}
    \vspace{4pt}
    \end{subfigure}
    \begin{subfigure}[t]{\textwidth}
    \includegraphics[width=\textwidth]{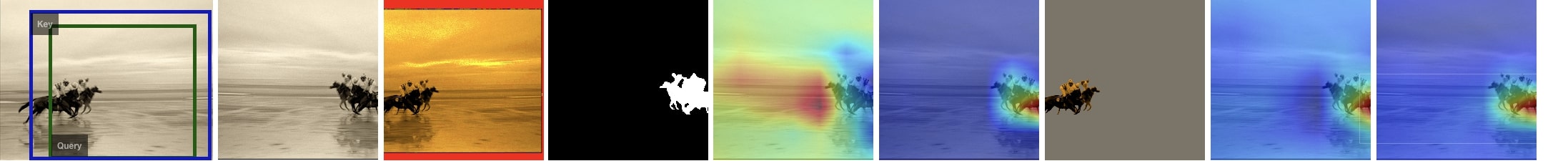}
    \vspace{4pt}
    \end{subfigure}
    \begin{subfigure}[t]{\textwidth}
    \includegraphics[width=\textwidth]{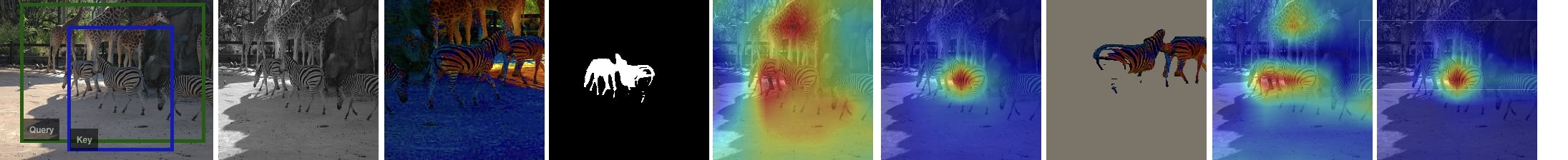}
    \vspace{4pt}
    \end{subfigure}
    \begin{subfigure}[t]{\textwidth}
    \includegraphics[width=\textwidth]{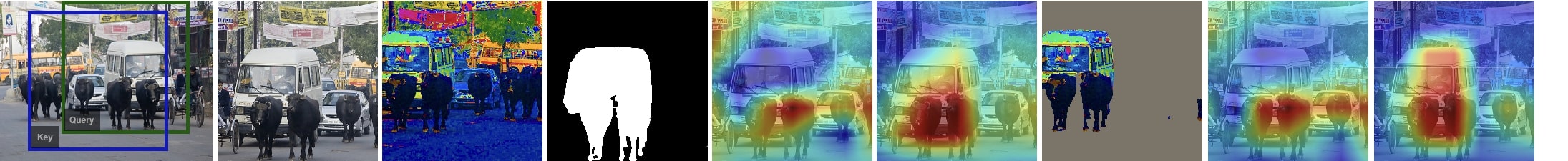}
    \vspace{4pt}
    \end{subfigure}
    \begin{subfigure}[t]{\textwidth}
    \includegraphics[width=\textwidth]{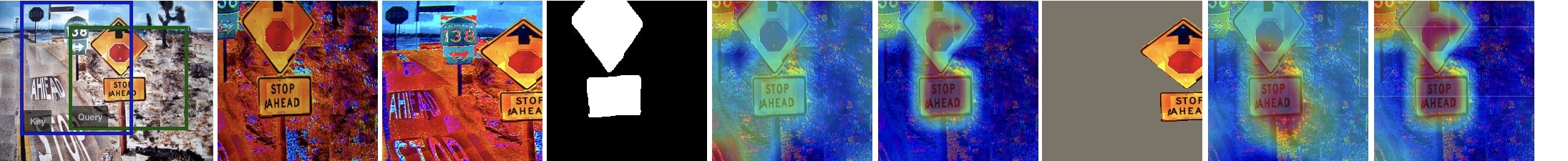}
    \vspace{4pt}
    \end{subfigure}
    \begin{subfigure}[t]{\textwidth}
    \includegraphics[width=\textwidth]{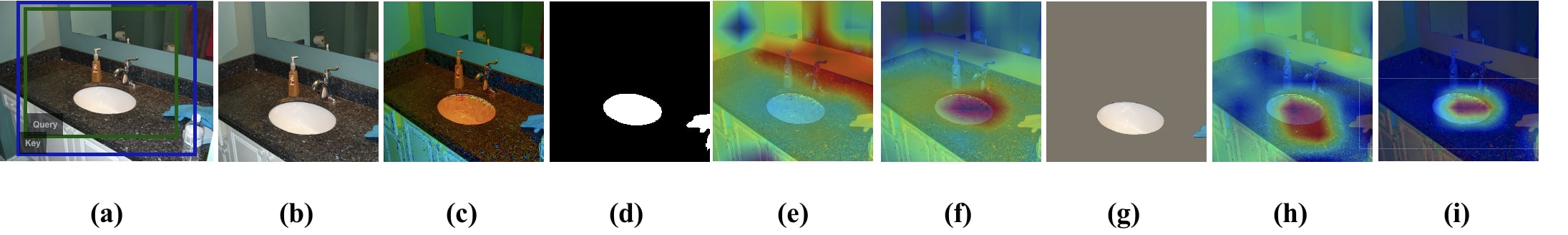}
    \vspace{4pt}
    \end{subfigure}
    
    \caption{CAST improves visual grounding of the contrastive  self-supervised feature encoder. 
    Column (d) shows the saliency map according to the query crop (b). Grad-CAM visualizations in columns (e, f) show the query regions that MoCo and MoCo + CAST models rely on, in order to match the key crop (c). Finally, the MoCo and MoCo + CAST models rely on query regions (h) and (i) to match with the masked key representation (g). Notice how in all the examples, models trained with CAST look at the salient objects of interest to match the query with the key or the saliency-masked key. 
    }
    \label{fig:sup_pretrained_grounding}
    \vspace{20pt}
\end{figure*}

Similar to Section 4.4 in the main paper, we use Grad-CAM to evaluate the visual grounding ability of a contrastive SSL model trained with CAST and its effect on grounding in downstream tasks. 
Randomly sampled examples in Figure~\ref{fig:sup_pretrained_grounding} show that the CAST-trained model seems to learn semantic category-specific feature representations, which allows it to look at objects of interest while performing query-key matching, and avoid learning spurious correlations. 

\subsection{Grounding of CASTed models on Downstream tasks}

\begin{figure*}[t]
    \begin{subfigure}[t]{0.5\textwidth}
    \includegraphics[width=\textwidth]{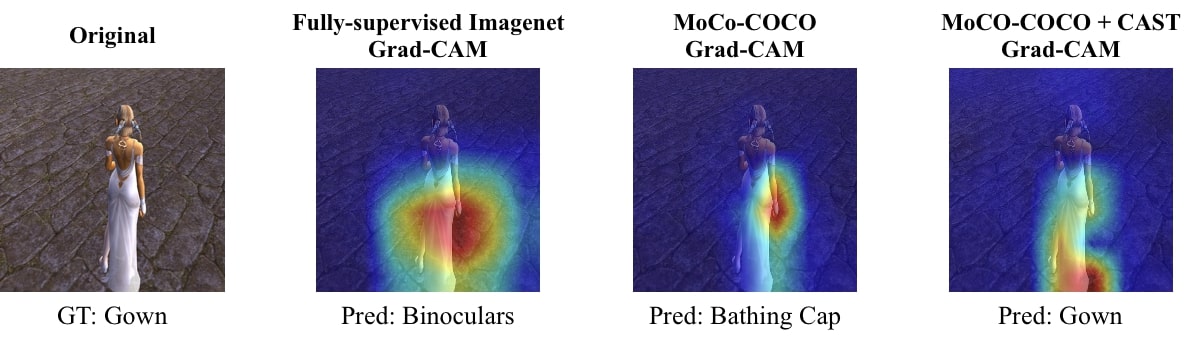}
    \caption{}
    \end{subfigure}%
    \begin{subfigure}[t]{0.5\textwidth}
    \includegraphics[width=\textwidth]{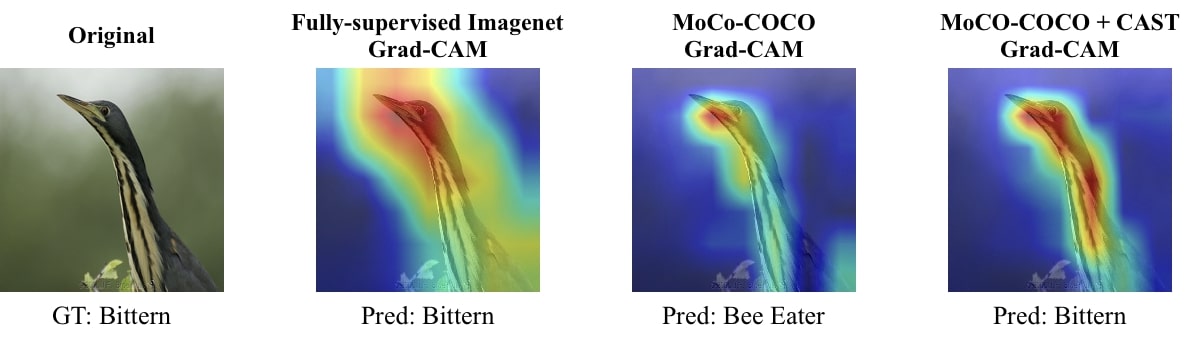}
    \caption{}
    \end{subfigure}
    \begin{subfigure}[t]{0.5\textwidth}
    \includegraphics[width=\textwidth]{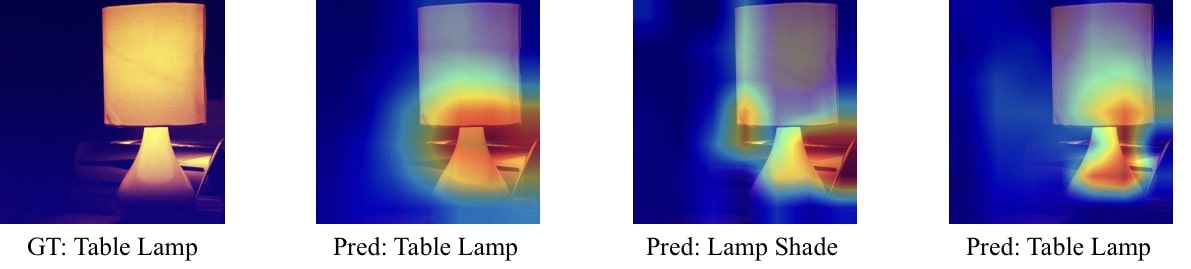}
    \caption{}
    \end{subfigure}%
    \begin{subfigure}[t]{0.5\textwidth}
    \includegraphics[width=\textwidth]{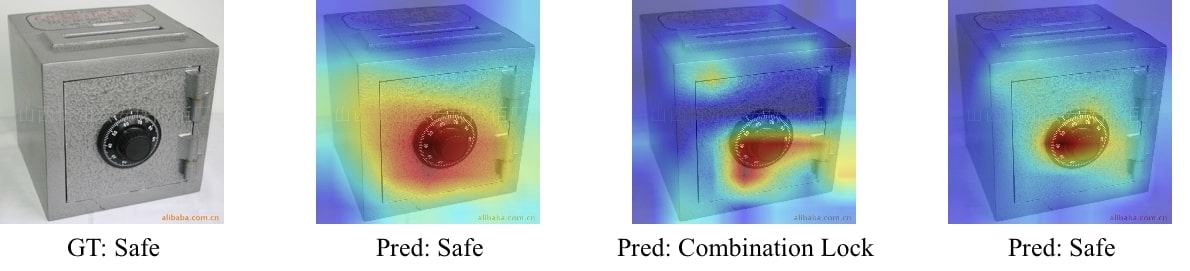}
    \caption{}
    \end{subfigure}
    \begin{subfigure}[t]{0.5\textwidth}
    \includegraphics[width=\textwidth]{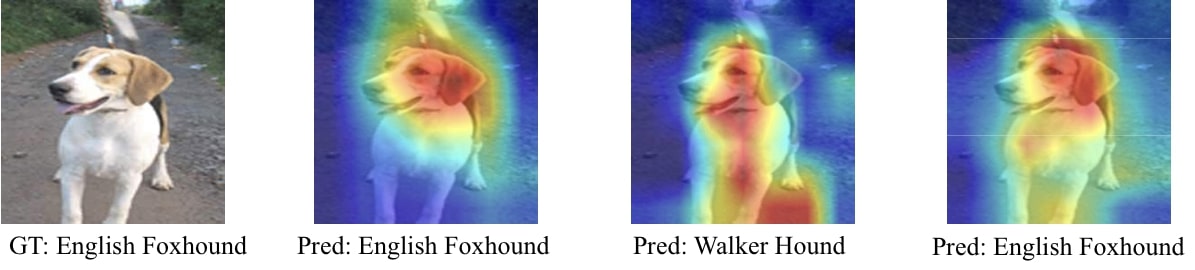}
    \caption{}
    \end{subfigure}%
    \begin{subfigure}[t]{0.5\textwidth}
    \includegraphics[width=\textwidth]{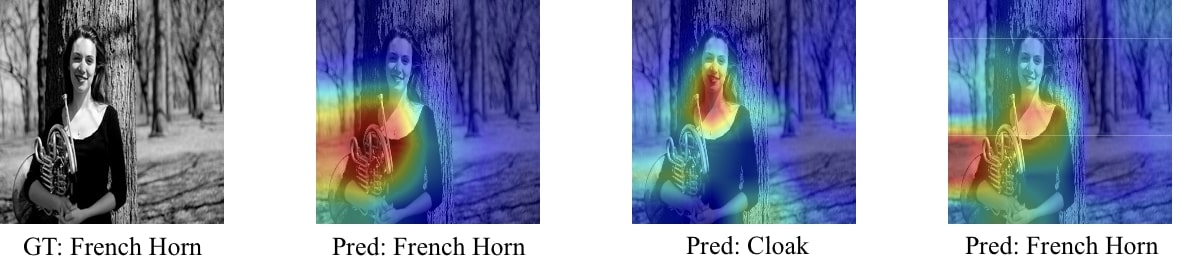}
    \caption{}
    \end{subfigure}
    \caption{Downstream task Grounding. Qualitative comparison of Grad-CAM attention maps for the correct class from fully supervised networks and the self-supervised networks (MoCo and MoCo + CAST) on the Imagenet-1k classification task showing cases where the CAST fixes the mistakes made by MoCo model.  We can see that CASTed models tend to look at the full extent of the object and less at the background regions. Note that even for incorrect predictions, Grad-CAM maps are computed for the ground-truth class. }
    \label{fig:supp_imagenet_grounding}
    \vspace{-5pt}
\end{figure*}

\reffig{fig:supp_imagenet_grounding} shows randomly sampled qualitative examples showing how training with CAST also leads to improvements in grounding in downstream tasks. 
Notice in \reffig{fig:supp_imagenet_grounding} (a, b) how the CASTed model looks at the whole extent of the object regions. 
We can also see in \reffig{fig:supp_imagenet_grounding} (c) how in order to predict ``Table Lamp", the MoCo pretrained model also looks at the table where as the CASTed model only looks at the relevant lamp regions. 
In \reffig{fig:supp_imagenet_grounding} (f), in addition to not predicting the correct class ``French Horn", the MoCo pretrained model is unable to localize the category, where as the CASTed model correctly predicts and attends to the ``French Horn". 

\subsection{Qualitative examples from Background Challenge}

\begin{figure*}[t]
    \begin{subfigure}[t]{\textwidth}
    \includegraphics[width=0.9\textwidth]{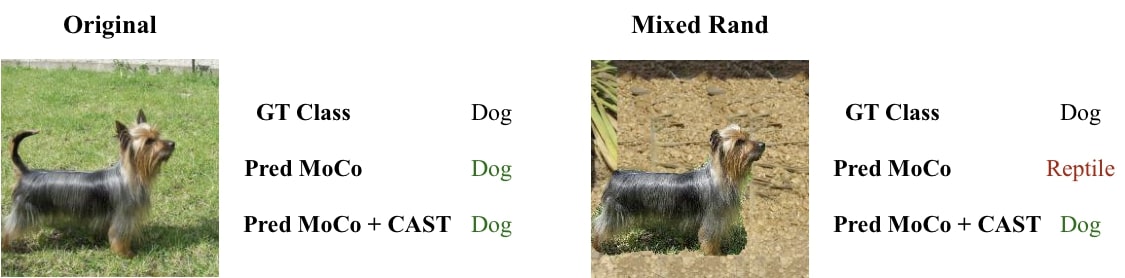}
    \caption{}
    \end{subfigure}
    \begin{subfigure}[t]{\textwidth}
    \includegraphics[width=0.9\textwidth]{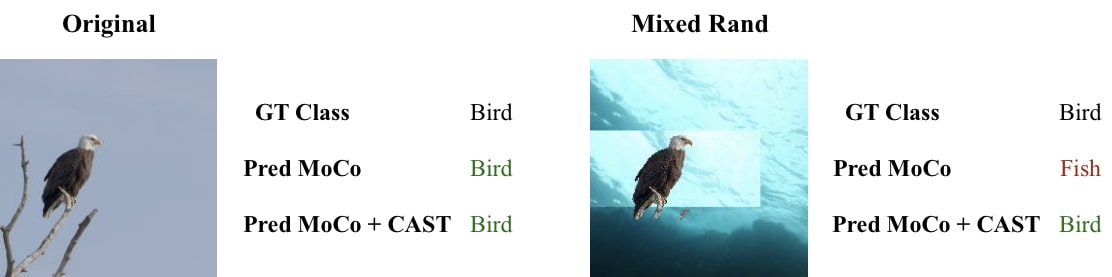}
    \caption{}
    \end{subfigure}
    \begin{subfigure}[t]{\textwidth}
    \includegraphics[width=0.9\textwidth]{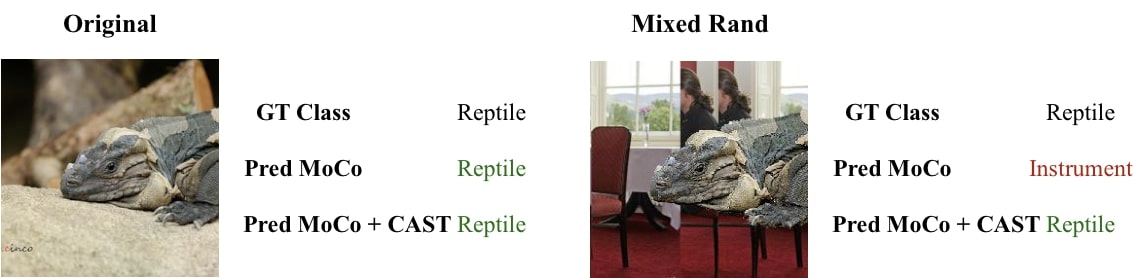}
    \caption{}
    \end{subfigure}
    \begin{subfigure}[t]{\textwidth}
    \includegraphics[width=0.9\textwidth]{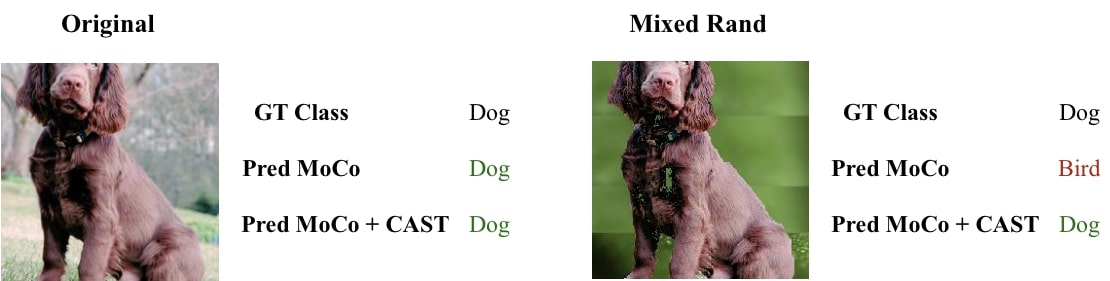}
    \caption{}
    \end{subfigure}
    \caption{Backgrounds-challenge. Models pretrained with CAST are less likely to rely on background correlations for classification. Notice how in (a), the MoCo pretrained model incorrectly predicts ``Reptile'' when the grass is changed to a background from a reptile class, while the CAST pretrained model still correctly predicts ``Dog''. Similarly in (c), when chairs are introduced in the background, the MoCo pretrained model changes its prediction from ``Reptile'' to ``Instrument'', indicating that the model relies on spurious correlations to predict the class of interest. The CAST pretrained model still correctly predicts ``Reptile.'' }
    \label{fig:sup_backgrounds_challenge}
    \vspace{15pt}
\end{figure*}

The Backgrounds Challenge~\cite{xiao2020noise} (\refsec{sec:backgrounds_challenge} of the main text) aims to assess the background-robustness of image classification models by measuring their accuracy on images containing foreground objects superimposed on various background types (see ~\cite{xiao2020noise} for details on dataset construction). 
Since CAST forces a model to attend to salient objects during learning, training with CAST also leads to models learning less spurious correlations, as seen in qualitative examples  (\reffig{fig:sup_backgrounds_challenge}). For example, in \reffig{fig:sup_backgrounds_challenge} (b), when the background corresponding to the sky is replaced by water, the MoCo pretrained model changes its decision from ``Bird'' to ``Fish'' indicating that the model was using spurious correlations associating water with fish. 
However the model pretrained with CAST, still correctly predicts the ``Bird'' class,  indicating that it relies more on the foreground object of interest while making decisions.

\end{appendices}

{\small
\bibliographystyle{ieeetr_fullname}
\bibliography{references}
}

\end{document}